\documentclass[journal]{vgtc}                     % final (journal style)
\vgtccategory{Research}

\vgtcpapertype{Analytics \& Decisions}

\title{Evaluating Machine Learning Models with NERO:\\Non-Equivariance Revealed on Orbits}

\author{%
  \authororcid{Zhuokai Zhao}{0000-0001-8201-2977},
  Takumi Matsuzawa,
  William Irvine,
  Michael Maire,
  and Gordon L Kindlmann
}

\authorfooter{
  \item Zhuokai Zhao, Michael Maire, and Gordon L Kindlmann are with Department of Computer
  Science, University of Chicago.
  	    E-mails: \{zhuokai, mmaire, glk\}@uchicago.edu.
  \item Takumi Matsuzawa and William Irvine are with James Franck Institute and Department of
  Physics, University of Chicago.
  	    E-mails: \{tmatsuzawa, wtmirvine\}@uchicago.edu.
}

\abstract{%
  Proper evaluations are crucial for better understanding, troubleshooting, interpreting model
  behaviors and further improving model performance.
  While using scalar-based error metrics provides a fast way to overview model performance, they
  are often too abstract to display certain weak spots and lack information regarding important
  model properties, such as robustness.
  This not only hinders machine learning models from being more interpretable and gaining trust,
  but also can be misleading to both model developers and users.
  Additionally, conventional evaluation procedures often leave researchers unclear about \textit
  {where} and \textit{how} model fails, which complicates model comparisons and further
  developments.
  To address these issues, we propose a novel evaluation workflow, named
  \textit{Non-Equivariance Revealed on Orbits (NERO) Evaluation}.
  The goal of NERO evaluation is to turn focus from traditional scalar-based metrics onto
  evaluating and visualizing models equivariance, closely capturing model robustness, as well
  as to allow researchers quickly investigating interesting or unexpected model behaviors.
  NERO evaluation is consist of a task-agnostic interactive interface and a set of
  visualizations, called NERO plots, which reveals the equivariance property of the model.
  Case studies on how NERO evaluation can be applied to multiple research areas, including
  2D digit recognition, object detection, particle image velocimetry (PIV), and 3D point cloud
  classification, demonstrate that NERO evaluation can quickly illustrate different model
  equivariance, and effectively explain model behaviors through interactive visualizations
  of the model outputs.
  In addition, we propose \textit{consensus}, an alternative to ground truths, to be used in NERO
  evaluation so that model equivariance can still be evaluated with new, unlabeled datasets.
}

\keywords{Interactive visualization systems and tools, explainable machine learning,
equivariance, integrated workflows.}

\teaser{
  \centering
  \includegraphics[width=\textwidth]{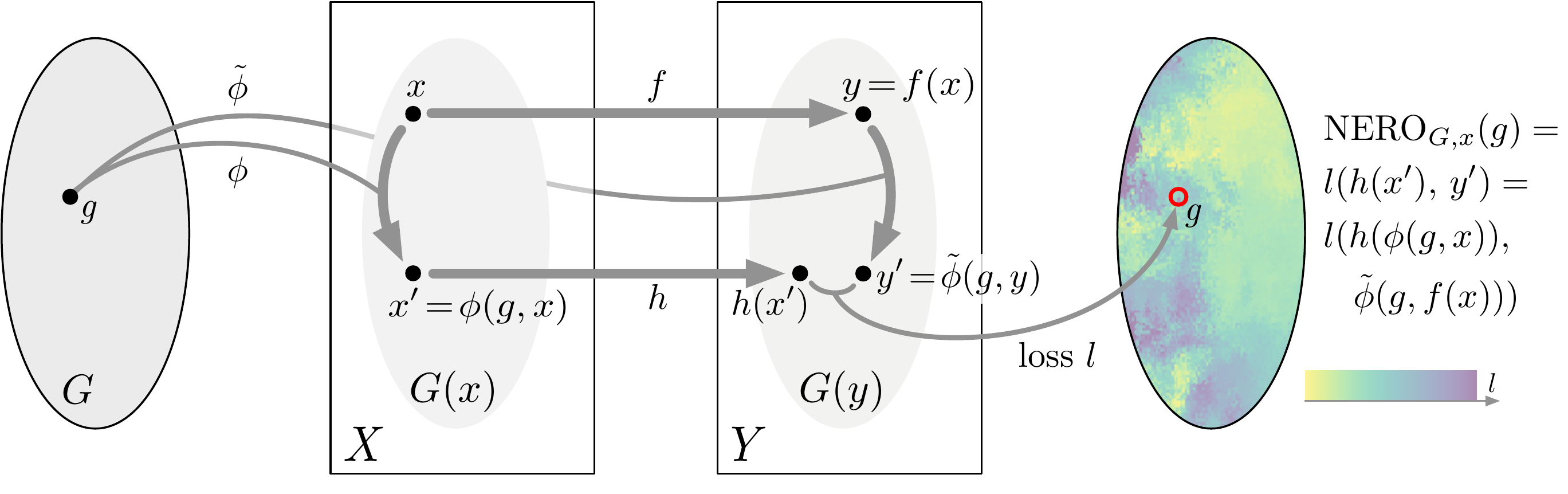}
  \caption{
  An ML model has inputs $X$ and outputs $Y$.
  $G$ is a transformation group acting on $X$ with $\phi$, and on $Y$ with $\tilde{\phi}$.
  The group element $g \in G$ transforms $x$ to $x' = \phi(g,x)$; the set of all
  possible transforms is the orbit $G(x)$.
  The model applied to $x$ is $\hat{y} = h(x)$, though this is not used in our method.
  Rather, the ground truth $y \in Y$ is transformed by $g$ to $\tilde{\phi}(g,y)$,
  which serves as ground truth to evaluate (here with loss function $l$) the result $h(x')$ of
  evaluating the model on transformed input $x'$.
  The $\mathrm{NERO}_{G,x}$ plot visualizes loss over the orbit $G(x)$.
  }
  \label{fig:general_illustration}
}

\renewcommand{\manuscriptnotetxt}{}

\nocopyrightspace

\graphicspath{{figs/}{figures/}{pictures/}{images/}{./}} % where to search for the images

\usepackage{tabu}                      % only used for the table example
\usepackage{booktabs}                  % only used for the table example
\usepackage{lipsum}                    % used to generate placeholder text
\usepackage{mwe}                       % used to generate placeholder figures

\usepackage{mathptmx}                  % use matching math font
\usepackage{caption}
\usepackage{subcaption}
\usepackage{amsfonts}

\begin{document}

%%%%%%%%%%%%%%%%%%%%%%%%%%%%%%%%%%%%%%%%%%%%%%%%%%%%%%%%%%%%%%%%
%%%%%%%%%%%%%%%%%%%%%% START OF THE PAPER %%%%%%%%%%%%%%%%%%%%%%
%%%%%%%%%%%%%%%%%%%%%%%%%%%%%%%%%%%%%%%%%%%%%%%%%%%%%%%%%%%%%%%%

\maketitle
%-------------------------------------------------------------------------
%% \section{Introduction} %for journal use above \firstsection{..} instead
\section{Introduction}\label{sec:introduction}
Applications of machine learning (ML), and deep learning (DL) more specifically, have enhanced and
accelerated many research areas, such as computer
vision~\cite{dos2021ViT,feich2022MAE,wang2022git} and natural language
processing~\cite{devlin2018bert,brown2020GPT,chowdhery2022palm}.
Thorough evaluation of ML models may deepen understanding of their functioning, and drive further
improvements.
The evaluation process in ML, unfortunately, remains largely unchanged over the past decade, which
hinders interpretation and innovation.

\begin{figure}
  \centering
  \includegraphics[width=\linewidth]{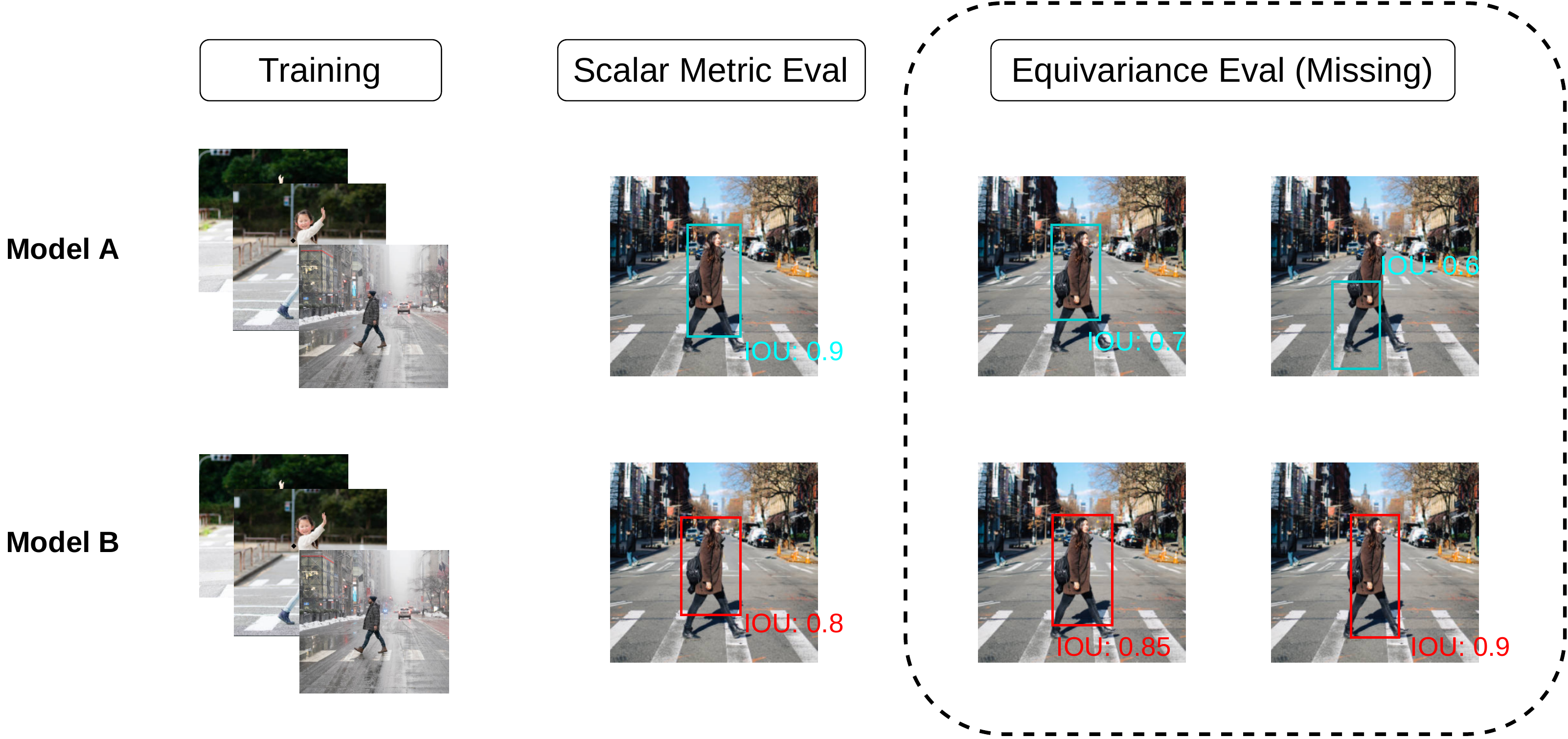}
  \caption{
    An example of scalar metric being ambiguous and misleading.
    Suppose object detection model $A$ and $B$ have been trained on the same dataset to detect
    human that is crossing the street.
    With standard evaluation procedure, both models are tested with some images and compared via
    Intersection Over Union (IOU).
    Model $A$ does a slightly better job than $B$.
    However, current result fails to characterize models in a more complete way.
    As shown in the dotted box, model $A$ might perform worse in corner yet important cases
    where the person is at the edge of the image, in which model $B$ has an advantage.
  }
  \label{fig:introduction_figure}
  \vspace{-0.1in}
\end{figure}

Model quality is typically measured with a scalar, such as accuracy for classification tasks,
precision and recall for object detection, and mean squared error for more quantitative tasks.
Though straight-forward, comparing models via scalar metrics can miss important details, limiting
insight for ML researchers, and creating ambiguities for practitioners.
Two models can be quantitatively similar on average, but respond very differently
to meaningfully changing individual inputs.
Fig.~\ref{fig:introduction_figure} illustrates two models trained to recognize
humans crossing streets.
A model that responds erratically to translating the field of view (which should only translate
the predicted bounding box) may be less trustworthy than one that performs better {\em on average}
on a fixed test set.

Empirical science provides especially challenging areas for applied ML, for at least two reasons.
First, specialized instrumentation means data is expensive to gather and labor-intensive
to label.
Popular ML models, however, need large training datasets, in part due to the simplicity of their
scalar metric evaluations.
The gold-standard dataset for image object detection,
Microsoft COCO~\cite{linMicrosoftCOCOCommon2015}, has $328,000$ labeled
images, and the more recent Object365~\cite{Object365} has over $2$ million.
The ubiquity of ML for object detection justifies and amortizes the cost of creating
such datasets, but this scaling does not generally apply to experimental science.
Second, scientists in particular value robustness, predictability and understandability in their
computational tools~\cite{Oviedo_2022}, in contrast to the black-box nature of deep learning.
These issues have catalyzed research in interpretable machine learning
(IML)~\cite{doshi-velezRigorousScienceInterpretable2017}, which seeks to intelligibly reveal
ingredients of ML model predictions.

Our work complements IML research by revealing how ML models respond to changing inputs, in a way
that is intuitive but mathematically grounded.
We focus on \textit{equivariance}, which captures how changes in model inputs map to changes in
outputs.
In Fig.~\ref{fig:introduction_figure}, for example, translating the input image should
consistently correspond to translations of the output bounding box.
We organize our visualization of model equivariance around a mathematical \textit{group} of input
transformations and the set of all transformations (the \textit{orbit}) of a given input.
This is captured in our proposed \textit{Non-Equivariance Revealed on Orbits (NERO) Evaluation},
which shows how equivariant a model is, and the structure of its equivariance failures.
In settings where practitioners can reason about their analysis task in terms of mathematically
predictable responses to data transforms, NERO evaluation gives an informative and detailed picture
of ML model performance that is missing from prior scalar summary metrics.

The contributions of this paper, are:
\begin{enumerate}
  \item \textit{NERO Evaluation}, an integrated workflow that visualizes model equivariance in an
  interactive interface to facilitate ML model testing, troubleshooting, evaluation, comparison,
  and to provide better interpretation of model behaviors,
  % \item the individual NERO plot as a way to visualize a model's equivariance on a given
  %   individual input,
  % %
  % \item aggregate NERO plots to visualize equivariance over a set of inputs, and dimension
  %   reduction plots to show the variety of equivariance failures in that set,
  % %
  % \item an interactive interface combining these plots in linked views,
  % %
  \item and \textit{consensus}, a proxy for ground truth that helps evaluate model equivariance
  with unlabeled data.
\end{enumerate}
%

%-------------------------------------------------------------------------
\section{Related Work}\label{sec:related_work}
% % Related work outline
% \noindent\fbox{%
%     \parbox{\linewidth}{%
%         Outline of this section (to be removed in actual paper):
%         \begin{itemize}
%             \item Related ENN work. Mainly using scalar, which has gap to be filled by NERO plots.
%             \item Related IML work and how it influences our work (vis related)
%         \end{itemize}
%     }%
% }
%
% \noindent Existing work from ENN and IML inspire the idea of developing NERO plots.
% %
% The great amount of efforts that have been put into improving equivariance for
% neural network models draws our attention.
% %
% While focusing on developing stronger equivariance as ML researchers do is
% undoubtedly important, we look at equivariance from a different angle and think it
% actually could also be an untapped great metric for better evaluation and
% interpretation of deep learning models.
% %
%
% Existing IML work, on the other hand, provides great examples on visualizing
% various neural network aspects, including model components, sensitivity and
% feature importance.
% %
% Therefore, NERO plots are built upon knowledge from both fields to visualize model
% equivariance.
% %
% In other words, NERO plots focus on representing model equivariance, a key concept
% in ML, via visualization.
% %
% Next, we explain in more details how existing work from ENN and IML connect with
% NERO plots.
% %
\subsection{Equivariant Neural Networks (ENNs)}\label{subsec:enn}
Improving model equivariance has become a popular ML research topic.
Models that are more equivariant have better generalization
capability~\cite{weilerGeneralEquivariantSteerable2021}, an important goal
of applied ML research.
% improving equivariance enables deep learning models to perform well without being trained with an
% extensive large amount of data.
%
% Such efforts would value even more when it comes to areas like physical science, medicine, where
% data samples are very time-consuming to label even for people with specific domain knowledge.
%
Equivariance sometimes occurs naturally in neural
networks~\cite{olahNaturallyOccurringEquivariance2020},
but guaranteeing equivariance requires more dedicated efforts.
Various works focus on improving equivariance of convolutional neural networks (CNN)
with respect to rotations~\cite{cohenSteerableCNNs2016a, kondorClebschGordanNetsFully2018a,cohenGaugeEquivariantConvolutional2019a, weilerGeneralEquivariantSteerable2021},
% ~\cite{,marcosRotationEquivariantVector2017, , kondorGeneralizationEquivarianceConvolution2018a, weiler3DSteerableCNNs2018, , cohenGeneralTheoryEquivariant2020, finziGeneralizingConvolutionalNeural2020,
shifts~\cite{zhangMakingConvolutionalNetworks2019a, azulayWhyDeepConvolutional2019, engstromExploringLandscapeSpatial2019, kayhanTranslationInvarianceCNNs2020, chamanTrulyShiftinvariantConvolutional2021},
and scales~\cite{xuScaleInvariantConvolutionalNeural2014, kanazawaLocallyScaleInvariantConvolutional2014, worrallDeepScalespacesEquivariance2019, ghoshScaleSteerableFilters2019, sosnovikScaleEquivariantSteerableNetworks2020} through network
architectural designs.
%
% These transformation groups on which existing works focus, such as rotation,
% shift, and scale, are the most common types that data often possesses.
% %
% As will be seen later in Section~\ref{sec:method} and~\ref{sec:experiments}, we
% employ similar groups of transformations as examples when presenting NERO plots.
%
%
% Domain-specific work has also shown impressive results, with recent benchmarks in
% point cloud segmentation~\cite{wangEquivariantMapsHierarchical2020}
% and molecular engineering, which includes force field
% prediction~\cite{batznerSEEquivariantGraph2021}, energy
% estimation~\cite{klicperaDirectionalMessagePassing2020} and 3D structure
% generation~\cite{satorrasEquivariantNormalizingFlows2021}.
% Domain-specific work has also shown impressive results, with recent benchmarks in
% point cloud segmentation~\cite{wangEquivariantMapsHierarchical2020}
% and molecular engineering~\cite{klicperaDirectionalMessagePassing2020,batznerSEEquivariantGraph2021, satorrasEquivariantNormalizingFlows2021}.
%
Data augmentation during training is also effective for improving
equivariance~\cite{chenGroupTheoreticFrameworkData2020}, with examples in
generative models~\cite{antoniouDataAugmentationGenerative2018, haubergDreamingMoreData2016,mirzaConditionalGenerativeAdversarial2014, sixtRenderGANGeneratingRealistic2017b},
Bayesian methods~\cite{tranBayesianDataAugmentation2017}, and reinforcement
learning~\cite{cubukAutoAugmentLearningAugmentation2019, ratnerLearningComposeDomainSpecific2017}.
%
% %
% These methods usually apply invariant transformations on training dataset to
% achieve better model performance.
% %
% In NERO plots, on the other hand, it is the evaluation dataset on which
% transformations are applied.
% %

Existing work often implicitly assumes that more equivariant models will have lower
errors when tested on large datasets, due to the close relationship between equivariance and
robustness~\cite{engstromExploringLandscapeSpatial2019, lagraveIntroductionRobustMachine2021}.
In fact, equivariance is indeed a close proxy for model robustness, as both represent the model's
ability to perform well on new dataset with or without similar distribution as the training
dataset, which is why the works mentioned above apply the same evaluation process using
standard scalar metrics for equivariance evaluations.
But still, the absence of evaluations directly showing model equivariance hinders more accurate
understanding of model behaviors, which inspired our work on NERO evaluation.
%
%In fact, we believe that the proposed NERO plots could be used when evaluating new ENN models in
%the future.  -> moved to figure work
%

\subsection{Interpretable Machine Learning (IML)}\label{subsec:iml}
Deep neural networks (DNN) have achieved great success in a variety of applications
involving images, videos, and audio~\cite{lecunDeepLearning2015}.
However, advances in DNN research are generally more empirical than
theoretical~\cite{poggioTheoreticalIssuesDeep2020}.
DNN models thus still largely work as black boxes, limiting how practitioners
interpret and understand model predictions~\cite{benitezAreArtificialNeural1997, doshi-velezRigorousScienceInterpretable2017}.
%
%This raises more concerns when there is no easy way to look into this black box and understand why
%certain features are being selected or why certain predictions are being made.
%

IML research addresses this with methods based on various different strategies
% GLK removed molnarInterpretableMachineLearning2020 since its also specific to surrogate models, below
%Since it is hard to probe directly inside the models, existing IML methods take many alternative
%routes.
%
that can be broadly summarized as: model components, model
sensitivity, and surrogate
models~\cite{liptonMythosModelInterpretability2017, guidottiSurveyMethodsExplaining2018, viloneExplainableArtificialIntelligence2020, molnarInterpretableMachineLearning2020}.
Of the three, surrogate
models~\cite{JMLR:v10:gorissen09a,ribeiroWhyShouldTrust2016,ribeiroAnchorsHighPrecision}
are not described further here since they have little methodological connection to NERO evaluation.
Visualizations for IML seek to transform abstract data relationships into meaningful visual
representations~\cite{hohmanVisualAnalyticsDeep2018}.
Studies have shown that interactive visualization is a key aspect of sense-making when it comes
to combining visual analytics with ML systems, which shapes our designs in presenting NERO
evaluation through an interactive interface~\cite{chatzimparmpasSurveySurveysUse2020}.
%
% \fbox{any more text here about how this Vis for IML work relates to NERO?}

%
% GLK is not sure what these two sentences actually add
%Our work complements existing IML work, and loosely fits in the feature importance category,
%where we evaluate model equivariance through revealing model's responses to changes in its inputs.
%
%Thoughtful visualizations play a very important role in NERO plots as well: unlike conventional
%evaluations, whose results are usually single numbers, NERO plots present model performance
%evaluated on a set of transformed input samples, which bring challenges of presenting all the
%evaluation results in a straight-forward yet comprehensive way.
%

\noindent\textbf{Model Components.}
IML work focusing on model components visualize the internals of a neural network.
Abadi et al.~\cite{abadiTensorFlowLargeScaleMachine2016} developed dataflow graphs in
TensorFlow~\cite{abadiTensorFlowLargeScaleMachine2016}, which visualizes the types of computations
that happen within a model, and how data flows through these computations.
%
% It informs developers what the model is consist of (e.g., layers), but no deeper information can
% be obtained.
%
Following this work, Smilkov et
al.~\cite{smilkovDirectManipulationVisualizationDeep2017a} improved the dataflow graph by showing
weights sent between neurons using different colors, curves, line thickness, as well as feature
heatmaps.
Plots in our NERO evaluation interface do not similarly visualize model components, but do employ
similar visualization ingredients.
%

% In contrast to visualizing weights and neuron activations, several research
% groups~\cite{selvarajuGradCAMVisualExplanations2017, chattopadhyayGradCAMImprovedVisual2018, cashmanRNNbowVisualizingLearning2018}
% proposed to do it backwards and visualize the gradient flow during back-propagation.
% %
% More recent work~\cite{liVisualizingNeuralNetworks2020} explores further along this path by adding
% Grand Tour visualizations of the neural network responses with respect to time (e.g., how the
% model weights change over time during the training process).
%
Beyond static visualizations, Yosinski et al.~\cite{yosinskiUnderstandingNeuralNetworks2015}
designed interactive visualizations of learned convolutional filters in neural networks,
and Kahng et al.~\cite{kahngActiVisVisualExploration2017} designed interactive system ActiVis
for visualizing neural network responses to a subset of instances.
The ActiVis interface supports viewing neuron activations at both subset and instance level,
similar to our NERO interface, though
the underlying quantities visualized and the goals differ.
%

% In more recent years, works on comparing different models and layers have drawn more attention.
% %
% Hoyt et.al.~\cite{hoyt_tshe_2021} visualized how succeeding layers in a classification model
% separate different classes.
% %
% And Li et.al.~\cite{li_umap_2021} combined UMAP with Grand Tour to propose a new tool named UMAP
% Tour, to inspect and compare similarities between layers from different neural network models.
% %
% It is worth mentioning that while not visualizing any layers, NERO plots enjoy the advantage of
% distinguishing models' difference by design, since it visualizes each models' non-equivariance
% that appear almost never the same.
% %
% More on this in Section~\ref{sec:method} and~\ref{sec:experiments}.
%
% Empirical study that employs user feedback on which visualization of these
% components is more understandable to humans has also been
% explored~\cite{jeyakumarHowCanExplain2020}.
%

\noindent\textbf{Feature Importance.}
Instead of visualizing model components, other approaches show feature
importance by analyzing how model predictions change in response to changes in input data,
in a way that is agnostic to the ML model choice.
Friedman's Partial Dependent Plot (PDP)~\cite{friedmanGreedyFunctionApproximation2001}
reveals the relationship between model
predictions and one or two features by plotting the average change in model prediction
when varying the feature value. %  over marginal distributions.
%
% PDP effectively displays feature's influence on model prediction in some cases.
%
% However, it overlooks heterogeneous situations where positive and negative changes in predictions
% cancel each other out when computing the average.
%
% Also it can become misleading when features are strongly correlated.
%
% Furthermore, due to its high computational cost, only a small set of features, usually one or two,
% could be explored at a time.
%
Goldstein et al.~\cite{goldsteinPeekingBlackBox2014} built on this with Individual
Conditional Expectation (ICE) plots that show model prediction changes due to
changing features in individual data points, rather than the average.

Subsequent works visualize expected conditional
feature importance~\cite{casalicchioVisualizingFeatureImportance2019}, conduct sensitivity
analysis~\cite{strumbeljExplainingPredictionModels2014a}, and improve PDP with less
computation cost~\cite{apleyVisualizingEffectsPredictor2019}.
%
%In addition,
% Apley et al.~\cite{apleyVisualizingEffectsPredictor2019} proposed an alternative to PDP which does
% not assume feature independence and is much less computationally-expensive.
%
% In contrast, Casalicchio et al.~\cite{casalicchioVisualizingFeatureImportance2019} proposed
% partial importance (PI) and individual conditional importance (ICI) plots that visualize the
% expected conditional feature importance instead of expected conditional predictions.
%
% Besides various methods that drew inspirations from ICE plots, Strumbelj et
% al.~\cite{strumbeljExplainingPredictionModels2014a} proposed a sensitivity analysis method that is
% specifically for classification or regression models.
%
Lundberg et al.~\cite{lundbergUnifiedApproachInterpreting2017a} present SHapley Additive
exPlanations (SHAP) that assigns each feature an importance value to explain why a certain
prediction is made.
Zhang et.al.~\cite{zhang_hdmr_2021} derived a more robust, model-agnostic method
from high-dimensional representations to measure global feature importance, which
facilitates interpreting internal mechanisms of ML models.
While NERO evaluation also employs data transformation and a response-recording mechanism, and
is also model-agnostic, it does not visualize feature importance per se.
Instead, it collects model responses with respect to data transformed by group actions as a whole,
and supports visualizations at both aggregate (group) and instance levels.
%
%In other words, its use is not limited to any certain types of models, which makes it more
%versatile than existing works such as~\cite{strumbeljExplainingPredictionModels2014a}.
%

%\noindent\textbf{Surrogate Models.}
%
%Instead of explaining black-box model directly, these methods use a different and
%ostensibly more interpretable surrogate model to approximate the black-box model and its behavior~\cite{molnarInterpretableMachineLearning2020},
%with an approximation that can be global over all predictions~\cite{JMLR:v10:gorissen09a}
%or more local~\cite{ribeiroWhyShouldTrust2016, ribeiroAnchorsHighPrecision}.
%
%
% Ribeiro et al.~\cite{ribeiroWhyShouldTrust2016} proposed Local Interpretable Model-agnostic
% Explanations (LIME) that trains local surrogate models such as a decision tree using input/
% prediction pairs around the instance of interest obtained from the target model to then explain
% this certain instance.
% %
% The same authors later proposed Anchors~\cite{ribeiroAnchorsHighPrecision}, which deploys the same
% perturbation-based strategy as LIME but using scoped rules called anchors instead of surrogate
% models.
% %
% Besides the discussed three main categories above, example-based
% methods~\cite{wachterCounterfactualExplanationsOpening2018, mothilalExplainingMachineLearning2020}
% also provide great insights in utilizing surrogate models in IML.
%
%
%NERO plots are not methodologically related to surrogate models.

%-------------------------------------------------------------------------
\section{Mathematical Background}\label{sec:math_background}
\subsection{Group Action and Group Orbit}\label{subsec:group_action_orbit}

\begin{figure}
    \centering
    \includegraphics[width=\linewidth]{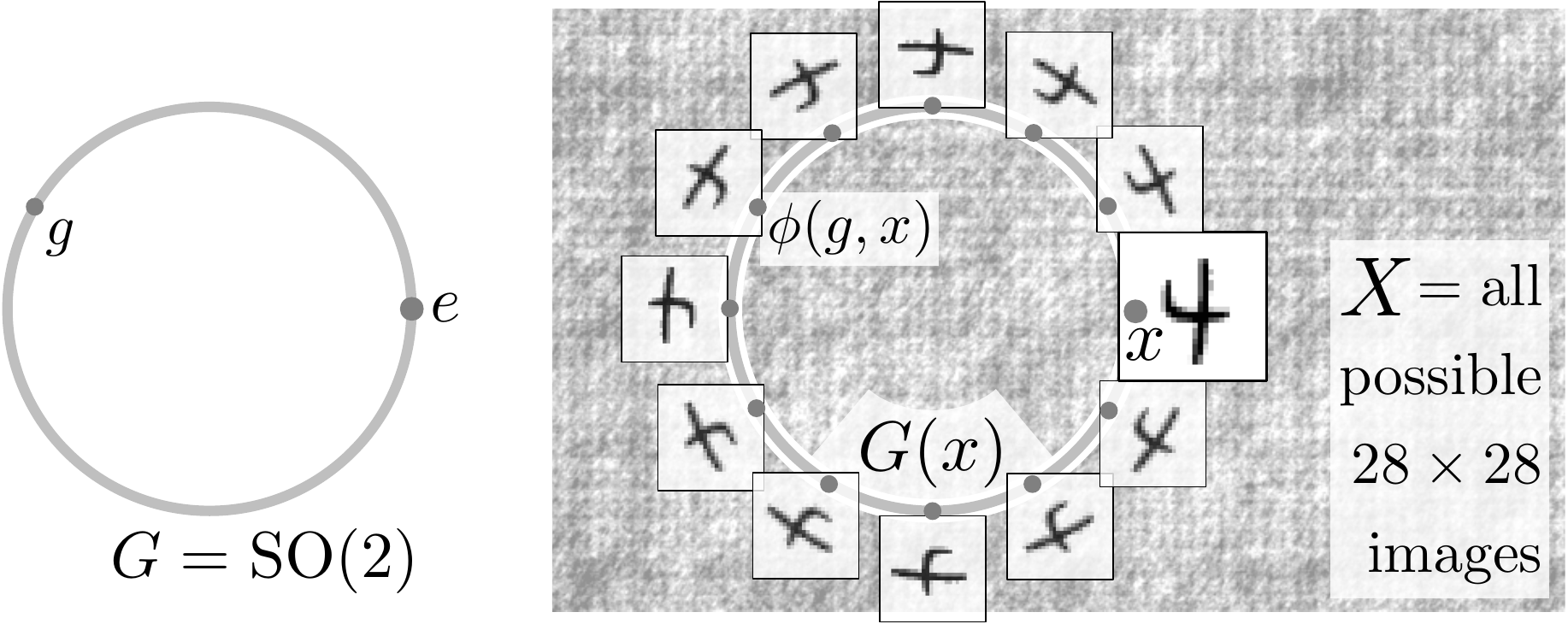}
    \caption{The group $G$ of 2-D rotations, left, acts on the set $X$ of images, right.
      An $x \in X$, a ``4'' digit, is rotated to $\phi(g,x)$
      for a $g \in G$ via group action $\phi$, part of the orbit $G(x) \subset X$ of
      all rotations of $x$.}
    \label{fig:group_action_diagram}
    \vspace{-0.1in}
\end{figure}

We give a concise summary of some elements of group theory, a rich topic meriting deeper consideration~\cite{rotman2012introduction}.
A {\em group} $G$ is a set with an operation ``$\cdot$''$: G \times G \rightarrow G $ that is
associative ($(f\cdot g) \cdot h = f\cdot (g \cdot h)$), with an identity element $e$ ($g \cdot e = e \cdot g = e$),
and with inverses ($g \cdot g^{-1} = g^{-1} \cdot g = e$).
A {\em group action} of group $G$ on set $X$ is a function $\phi : G \times X \rightarrow X$ that
transforms an $x \in X$ by $g,h \in G$ according to $\phi(g, \phi(h, x)) = \phi(g\cdot h, x)$
and $\phi(e, x) = x$.
The {\em orbit} of $x \in X$ under a group action $\phi$ is the set of all possible transformations
$G(x) = \{\phi(g,x) | g \in G\}$.
We use group orbits to generate a mathematically coherent family of ML model inputs,
with which (human) users of the model can predict and reason about corresponding model outputs.
For example, Fig.~\ref{fig:group_action_diagram} illustrates a single $28\times28$
MNIST~\cite{lecunGradientbasedLearningApplied1998} digit image $x$,
and its orbit under the rotation group $SO(2)$ through the space $X$ of all possible $28\times28$
images.
We currently make NERO plots for spatial transformation group actions (shifts, rotations,
flips), which have natural spatial layouts (e.g. the circular domain of $SO(2)$), but we want to
highlight that NERO plots should in principle work with any group that has an intelligible layout.

\subsection{Equivariance}\label{subsec:equivariance}
Three terms -- invariance, equivariance, covariance -- for describing the relationship between
changes in inputs and outputs of ML models~\cite{marcosRotationEquivariantVector2017}, can be
introduced via a commutative diagram~(\ref{eq:commdiag}).
\begin{equation}
 \raisebox{-3.5em}{\includegraphics[width=0.8\columnwidth,trim=0mm 0mm 0mm 0mm]{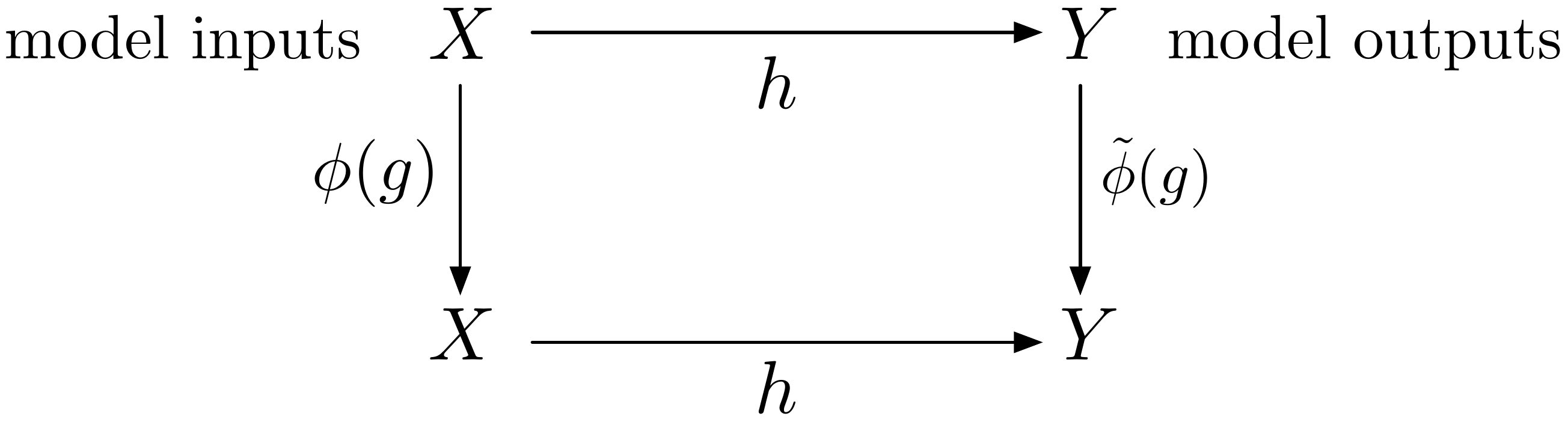}} \label{eq:commdiag}
\end{equation}
The ML model hypothesis $h$ maps from inputs $X$ to outputs $Y$.
For some group
element $g$, actions $\phi(g)$ and $\tilde{\phi}(g)$ transform $X$ and $Y$, respectively.
Assuming (\ref{eq:commdiag}) is true for some model
(i.e., hypothesis $h$ and transform $\tilde{\phi}(g)$ always reach the same output as input
transform $\phi(g)$ followed by $h$), the following definitions describe {\em how}.

The model is {\em invariant} with respect to the group action $\phi$
if $\tilde{\phi} = I$, the identity transform on $Y$.
In classification tasks, invariance means that the classification result
is unchanging while inputs are transformed in some way.
A model is {\em equivariant} when the model inputs and outputs are transformed in the same
way: $\phi = \tilde{\phi}$.
For example, in object detection, where model outputs are object bounding boxes,
if the object is shifted 5 pixels to the right, an equivariant model
would predict the bounding box 5 pixels to the right.
{\em Covariance} is an extension of equivariance in which $\phi$ and $\tilde{\phi}$ are
mathematically distinct (because $X$ and $Y$ have distinct types), but have a semantic
linkage necessitated by the structure of group $G$.
For example, in optical flow recovery from image sequences,
rotating the image inputs to a covariant model will produce an output in which both the
vector field domain and the vectors themselves are correspondingly rotated.
By a slight abuse of terminology, we use ``equivariance'' in this work to refer to all three
commutative diagram properties.
%

%------------------------------------------------------------------------
% NOTE: fig:general_illustration is in 0_main.tex (teaser image)
\section{Method}\label{sec:method}
\subsection{Overview}\label{subsec:methodover}
Diagram~(\ref{eq:commdiag}) describes an ideal, perfectly equivariant model.
Real models, applied to real data, often fall short of this; NERO evaluations seek to reveal how
through visualizations.
Fig.~\ref{fig:general_illustration} defines the NERO plot as inspired by
diagram~(\ref{eq:commdiag}): the thickest arrows at the center of the figure, within and between
$X$ and $Y$, roughly correspond to the arrows of (\ref{eq:commdiag}).
Input $x$, however, maps to ground truth $y$ rather than model output $h(x)$, and the purpose of
the NERO plot is to visualize the \textit{gap} between $h(x')$ and $y'$, where $h(x')$ is the model
output on transformed input $x'$, and $y' = \tilde{\phi}(g,y)$ is the transformed ground truth $y$.
This illustration uses an abstract depiction of group $G$ to schematically indicate $G(x)$ and
$G(y)$, but some particular spatial layout of $G$ necessarily determines the shape of the NERO
plot.
\textit{If the model is equivariant, then $h(x') = y'$, so the NERO plot is a flat constant.}
The visual structure of a non-constant NERO plot shows the structure of model non-equivariance over
the group orbit domain.

The quantity shown in a NERO plot is some scalar metric (understandable to practitioners) that
measures the gap between $h(x')$ and $y'$, including the standard metrics of model prediction
confidence, accuracy, mean square error (MSE) and more generally speaking, error metrics.
The NERO plot illustrated in Fig.~\ref{fig:general_illustration} (right) is an
\textit{individual NERO plot}, as it depicts model non-equivariance along the group orbit $G(x)$
around an individual input sample $x$.

While \S\ref{sec:introduction} critiqued single scalars to summarize model results over a large
dataset, informative NERO plots can also involve averaging.
An \textit{aggregate NERO plot} visualizes the average scalar metric over a dataset, or subset
of it, at each point along the group orbit (i.e. with the same domain as the individual NERO plot),
to show trends in the model's response to transformed inputs.
Like PDP and ICE plots (\S\ref{subsec:iml}), NERO plots evaluate the model within some
neighborhood around a given sample, but instead of varying features in isolation, we traverse the
orbit of some interpretable transform group.
\begin{figure*}
    \centering
    \includegraphics[width=.8\textwidth]{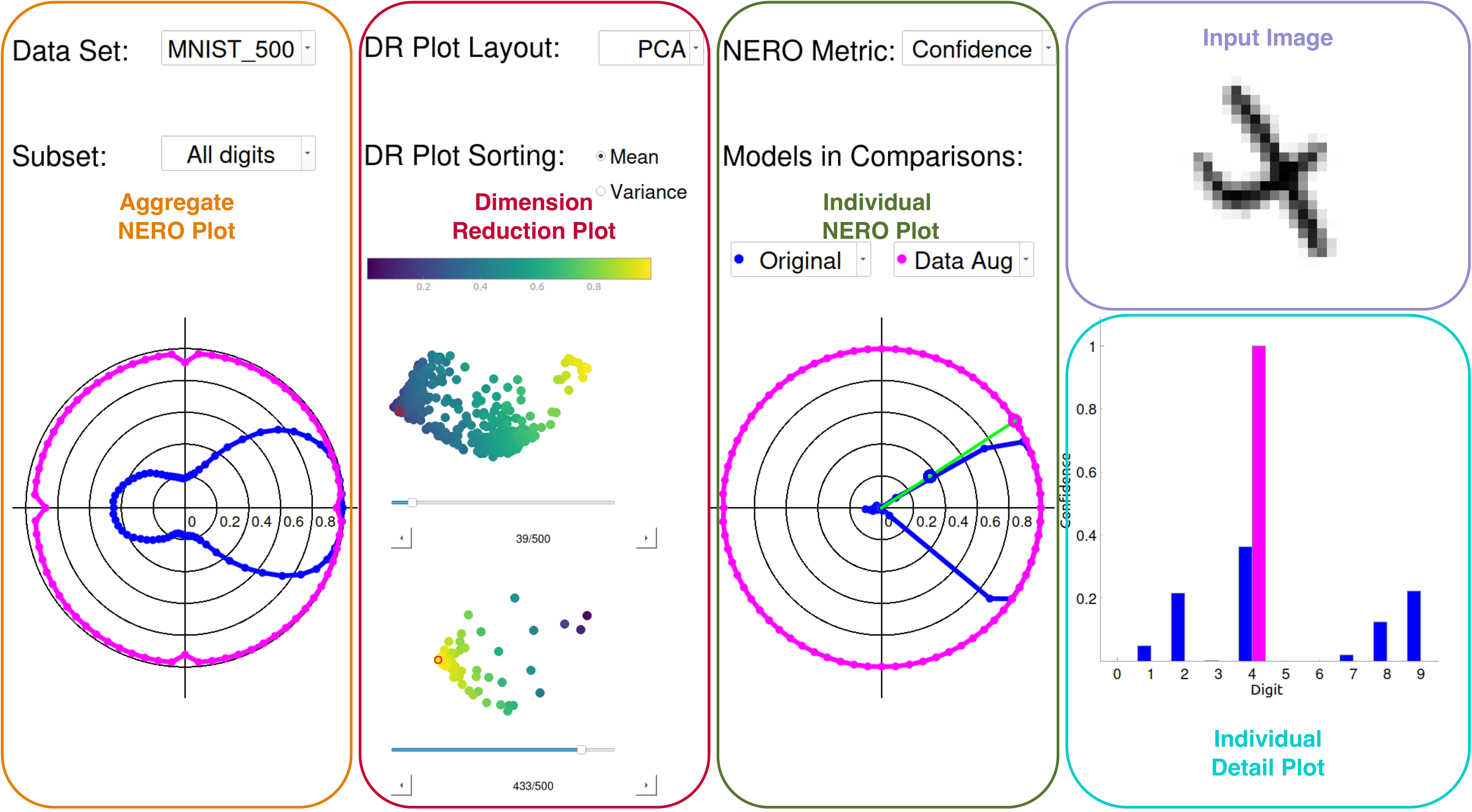}
    \caption{
      The NERO interface has $5$ sections: aggregate NERO plot, dimension reduction (DR)
      plot, individual NERO plot, input image, and detail plot.
      Each section is highlighted with orange, red, dark green, purple or cyan bound box
      respectively, where bounding boxes are not part of the interface design.
      The sections are interactively controlled, with linked views.
    }
    \label{fig:mnist_interface}
    \vspace{-0.1in}
\end{figure*}

To try to see degrees of freedom lost in the aggregate NERO plot, we can also
treat the individual NERO plots as $n$-vectors, and use dimensionality reduction.
The resulting \textit{dimension reduction (DR) scatterplot} organizes data points according to the
similarity of their patterns of non-equivariance, to help localize abnormal model behavior and
identify the connections between worse-performing cases.

All of these visualizations are linked together in the interactive \textit{NERO interface} that
provides users with both the convenience to see model equivariance in a high-level view
across a whole dataset (through the aggregate NERO plot), as well as navigating into detail views
(through the individual NERO plot, e.g., a specific place in the orbit where the model has
trouble).

\begin{figure}[ht]
   \begin{center}
       \includegraphics[width=\linewidth]{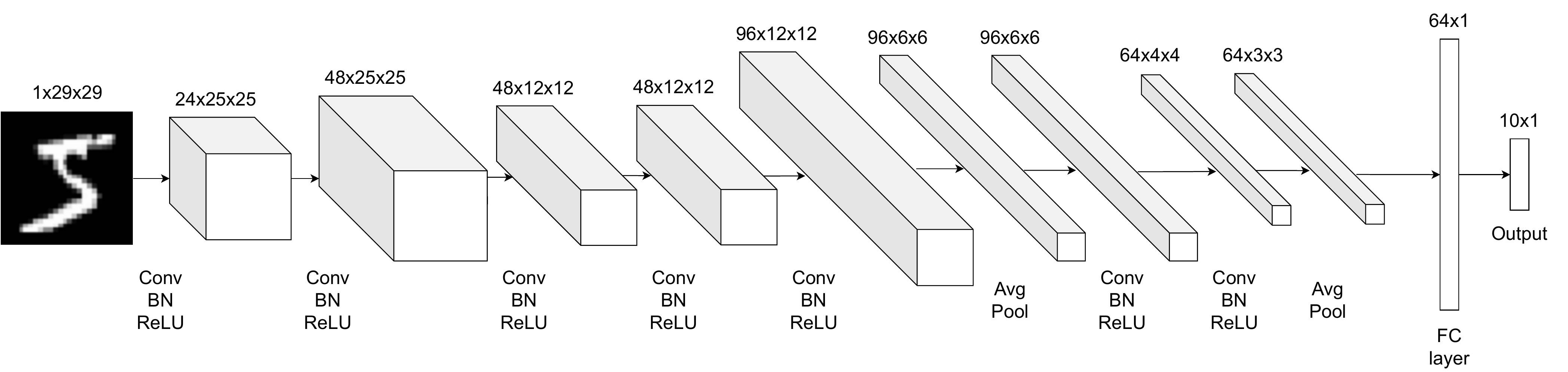}
       \caption{Network structure of the CNN model for digit recognition with MNIST~\cite{lecunGradientbasedLearningApplied1998}}
       \label{fig:mnist_network}
   \end{center}
   \vspace{-0.3in}
\end{figure}

The following subsections describe the components of NERO evaluation through a digit
recognition task with MNIST dataset~\cite{lecunGradientbasedLearningApplied1998}, with the group
action of continuous rotation around the image center..
NERO evaluation is presented via an interactive NERO interface, an example of which is in
Fig.~\ref{fig:mnist_interface}, starting with the individual NERO plot on the right,
The goal of using this task and MNIST dataset, is to utilize a well-known, easy-to-interpret
task as an example to make the illustrations of NERO evaluation more concretely, instead of
limiting or bounding NERO evaluation to only such or similar tasks, as we will be showcasing how
it could be applied to different use cases in \S\ref{sec:experiments}.
The CNN model here has six cascaded convolutional layers, six batch normalization (BN)
layers~\cite{ioffeBatchNormalizationAccelerating2015}, six rectified linear units
(ReLU)~\cite{glorotDeepSparseRectifier2011}, followed by two fully connected layers.
The detailed network architecture is illustrated in Fig.~\ref{fig:mnist_network}, but we would
like to point out that NERO evaluation is model-agnostic, and the purpose of explaining model
structure is to ensure reproducibility.

For the proposed NERO evaluation to be effective, the first criteria is to ensure that the
associated NERO plots are distinguishable enough when evaluated on two models with different
equivariance.
To illustrate how NERO plots differ on equivariant and non-equivariant models, the network
in Fig.~\ref{fig:mnist_network} is trained twice, first without and then with rotational data
augmentation, to create two models that predictably differ.
The augmentation model should have better invariance, even though the total amount of training
(with or without augmentation) is the same.

\subsection{Individual NERO Plot}\label{subsec:mnist_individual}
Individual NERO plots (Fig.~\ref{fig:mnist_interface} right) visualize model equivariance for a
single sample.
The NERO metric in this case is confidence: the probability of correct classification.
Individual NERO plot displays polar plots of confidence over the image rotation angle $\theta$,
for a particular input image of a ``4'' (Fig.~\ref{fig:mnist_interface} upper-right corner).
For a model with perfect rotational equivariance, the individual NERO plot will be a circle, while
any dips indicate non-equivariance.
The NERO plots for the \textit{original model} trained without data augmentation, and for the
\textit{DA model} trained with augmentation, are in blue and magenta, respectively.

The plots confirm our expectation that the DA model is more equivariant, with the magenta
plot being a near-perfect circle, while the blue (original model) plot is highest at
small rotation angles, which proves the plot being distinguishable between different models.
For a single interactively selected rotation angle (green line in polar plot), the details of the
models' predictions are shown as a bar chart (Fig.~\ref{fig:mnist_interface} lower-right corner)
showing confidences for all possible digits.
Such a detail view is necessarily specific to the model task and data type, but the NERO interface
should have any visualization of individual plots, input data sample, and model details to be
adjacent.
Here, the DA model (magenta) has higher confidence in recognizing ``4'' and essentially zero
confidence for any other digit, unlike the original model (blue), which is highest for ``4'' but
with non-zero confidence for other digits.
\begin{figure}
    \centering
    \includegraphics[width=\linewidth]{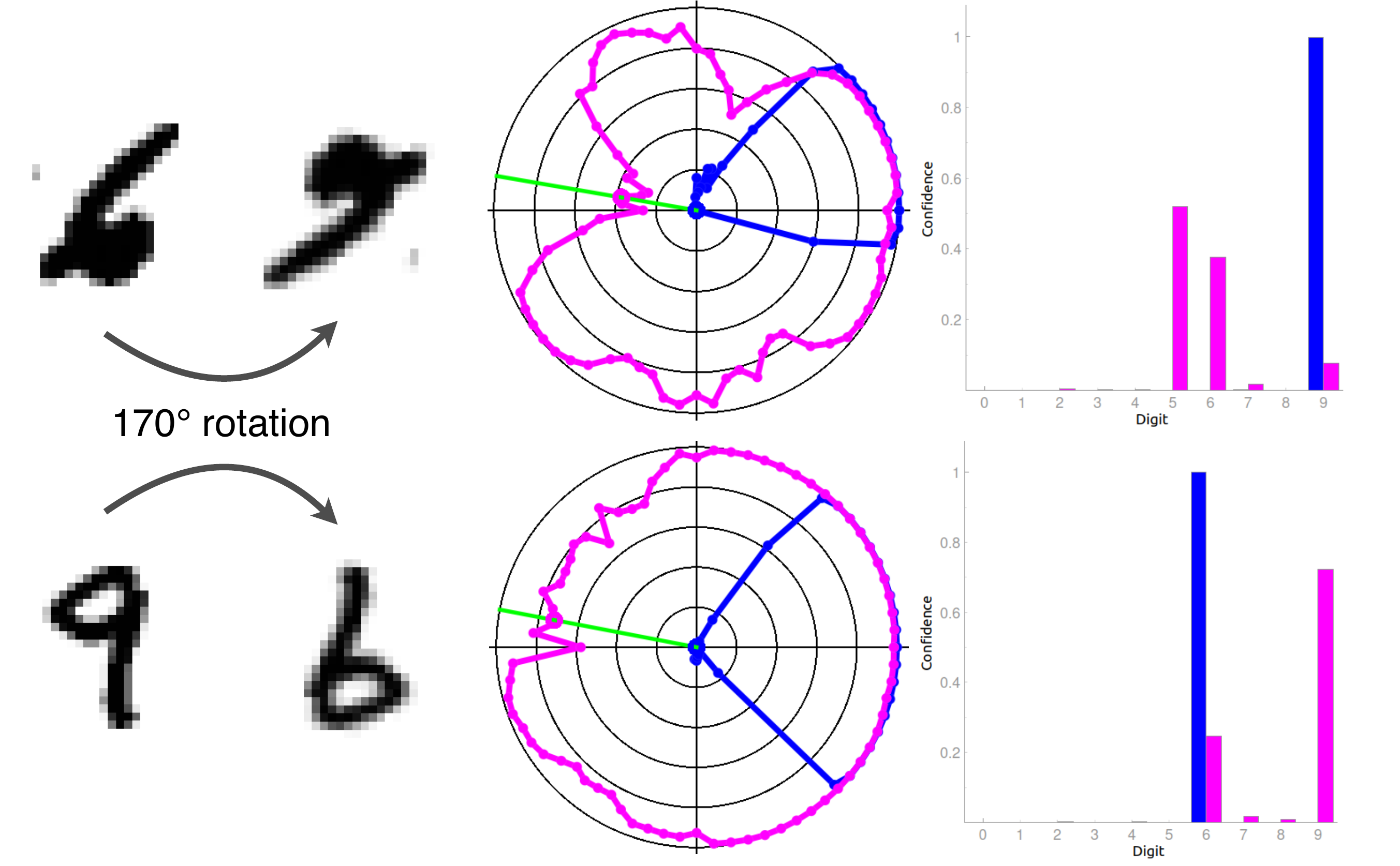}
    \caption{
      The individual NERO and detail plots of original (blue) and data augmented (magenta) models,
      for two digits rotated $170^\circ$, ``6'' (top) and ``9'' (bottom), reveal the extent to
      which data augmentation overcome the confusion between these two rotated digits.
    }
    \label{fig:mnist_individual_interactions}
    \vspace{-0.1in}
\end{figure}

Some insights about the structure of data and task can be gleaned from individual NERO plots,
for example, the digits $6$ and $9$ in Fig.~\ref{fig:mnist_individual_interactions}.
For both digits, the original and DA models perform similarly at zero or small rotations angles
while the original model fails as the angle increases (rightward lobe in blue plots), whereas the
DA model performs better (though not uniformly) over all angles (magenta plots).
The individual NERO plots show that the original model confidence falls to near zero for large
angles, but the detail bar charts (Fig.~\ref{fig:mnist_individual_interactions} right) provide
additional insight: the original model mis-classifies the $170^\circ$-rotated $6$ as $9$, and the
rotated $9$ as $6$, consistent with these digits' basic shapes.
The DA model does not have the same near perfect equivariance as with the ``4'' digit of
Fig.~\ref{fig:mnist_interface}, but the level of equivariance here is still surprising: the DA
model (magenta) gives moderate confidence of ``6'' for the rotated $6$, and highest confidence of
``9'' for the rotated $9$, with lower confidence for the incorrect digits.
The performance of the DA model implies that the shapes of $6$s and $9$s within MNIST are distinct
enough (9s having a straighter side) that they may be correctly recognized even with rotation.
This exemplifies how individual NERO plots with detail views can not only visualize model
equivariance on a single sample, but also help interpret model characteristics.
We will show in \S\ref{sec:experiments} how individual NERO plots and detail views help visualize
equivariance and provide interpretations for other models and tasks.

\subsection{Aggregate NERO Plot}\label{subsec:mnist_aggregate}
Aggregate NERO plots reveal over-all equivariance for a subset of a dataset, or an entire dataset,
using the same spatial orbit layout and the same visual encoding as in the individual plots,
though the scalar quantity visualized may be different.
The aggregate NERO plots on the left side of Fig.~\ref{fig:mnist_interface} show equivariance
for $500$ MNIST images, with $50$ images of each digit, using the same polar plots over the
circular domain of the rotation group orbit.
The aggregate plots, however, show \textit{accuracy} -- the fraction of correct classifications
over the input samples -- rather than the confidence shown in the individual plots.
%
% Because accuracy is a more commonly used metric for assessing model performance on larger
% datasets, and may be a more understandable metric for the aggregate plots.
%
In our MNIST example, the data augmented model (magenta) is much more equivariant for this
subset of images than the original model (blue).
\begin{figure}
    \centering
    \includegraphics[width=\linewidth]{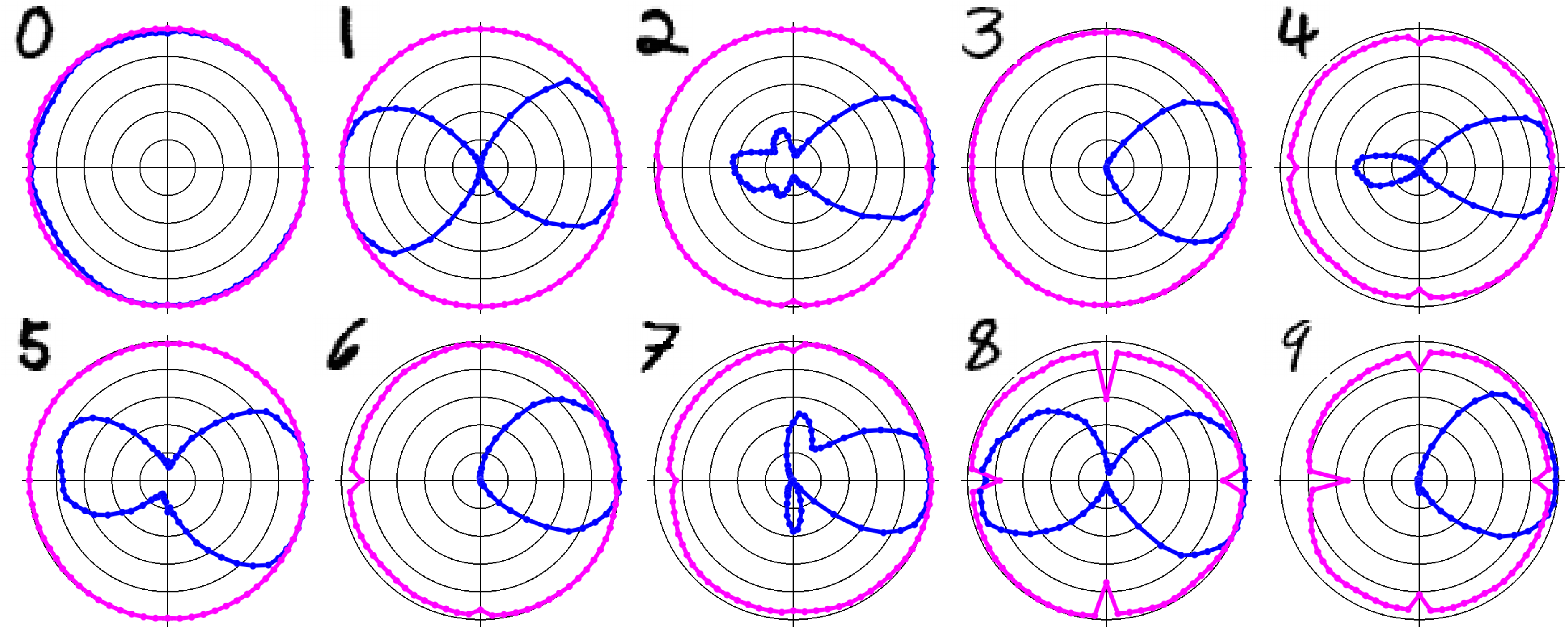}
    \caption{
      Aggregate NERO plots for the original (blue) model reflect the average rotational symmetry
      of each digit.
    }
    \label{fig:mnist_aggregate_interactions}
    \vspace{-0.1in}
\end{figure}

Aggregate NERO plots also reveal additional properties of the task and data.
Fig.~\ref{fig:mnist_aggregate_interactions} shows aggregate plots for $50$ images of each digit.
Digit $0$ is already rotational invariant, so both original and DA NERO plots show equivariance.
The original model (blue) aggregate NERO plot for ``1'' shows its $180^\circ$ rotational symmetry
with lobes at $0$ and $180$ degrees; the same holds for $8$ and to a lesser extent for $5$.
The lack of rotational symmetry of digits $2$, $3$, $4$, and $7$ are all confirmed by
their blue plots.
Even though NERO plots cannot answer questions about \textit{why} a model made the predictions
it did (as pursued in other interpretable machine learning work, \S\ref{subsec:iml}), these
examples suggest how aggregate NERO plots can be used to help understand patterns of
model behavior with specific input classes over specific transforms.

\subsection{Dimension Reduction (DR) Plot}\label{subsec:dr_plot}
Dimension reduction (DR) plots conceptually bridge the information in the aggregate and individual
NERO plots, and are thus shown in between them in the NERO interface (second part of
Fig.~\ref{fig:mnist_interface}).
The data vector underlying the individual NERO plot (all the metric values evaluated over the group
orbit) is considered as a point in some high-dimensional space, and a dimensionality reduction
method is applied to lay out the data points in a 2D scatterplot.
Our current interface supports layout via principle component analysis (PCA), independent
component analysis (ICA), as well as non-linear ISOMap, t-SNE~\cite{vanDerMaaten2008}, and
UMAP~\cite{umap_2018}; Fig.~\ref{fig:mnist_interface} shows results with PCA.
The intent is that data samples with similar patterns of non-equivariance should be nearby in the
DR plot, to give an over-all sense of the varieties of non-equivariance from that model, and to
highlight any outlier inputs requiring detailed attention.
The scatterplot dots are color-encoded by either the mean or the variance of the individual NERO
plot values; mean for showing trends in over-all model performance, and variance for showing which
inputs exhibited the best or worst equivariance.
\begin{figure}
    \centering
    \includegraphics[width=\linewidth]{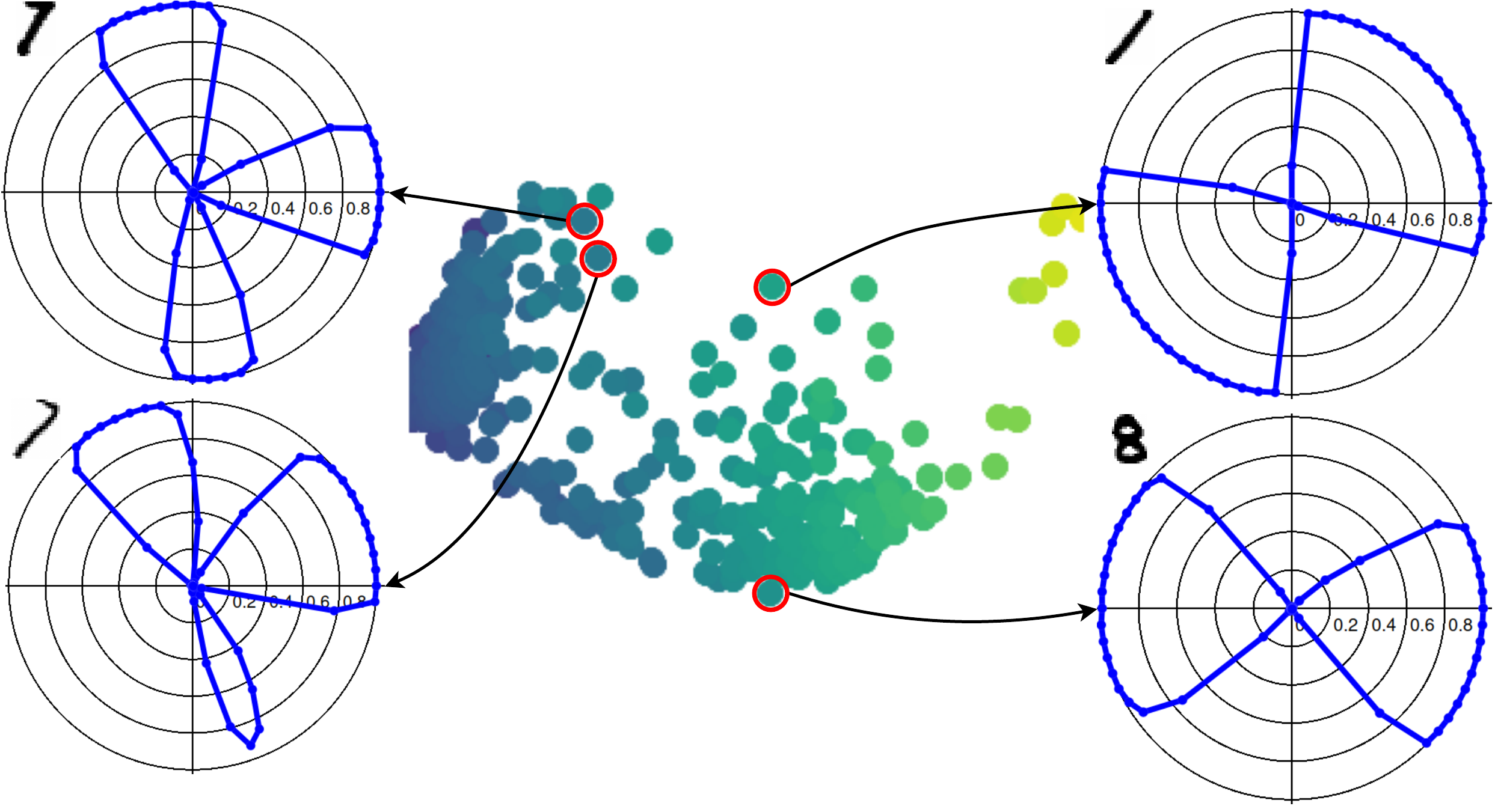}
    \caption{
      DR plot colormapped by mean confidence, annotated with some associated individual NERO plots.
    }
    \label{fig:mnist_dr_plot}
    \vspace{-0.1in}
\end{figure}

In our interactive interface, users can click on a dot of interest in the scatterplot to trigger
display (to the right) of the corresponding individual NERO and detail plots;
in Fig.~\ref{fig:mnist_interface} the selected point is indicated with a small red circle.
Fig.~\ref{fig:mnist_dr_plot} shows an expanded view of this scatterplot, annotated
with individual NERO plots for selected points.
These suggest that the DR plot is successful in presenting a navigable view of the different
patterns of non-equivariance, with similar individual plots (Fig.~\ref{fig:mnist_dr_plot} left)
arising from nearby points.
More distant points have distinct individual plots (Fig.~\ref{fig:mnist_dr_plot} right), though in
this case the similarly shaped plots are quantitatively distant because of their different
orientations.

\subsection{NERO Interface}\label{subsec:nero_interface}
% {\em (The NERO interface is demonstrated in the supplementary video.)}
%
The previously described components of the NERO interface (Fig.~\ref{fig:mnist_interface}) are
designed with the general logic of overview on the left and details on the right; this spatial
layout is the same across different applications, as will be shown in more case studies
in \S\ref{sec:experiments}.
All sections are individually controllable and interactively linked.
On the left, the dataset and subset of interest are selected via drop-down menus, with the
resulting aggregate NERO plot below.
The DR plot section supports choosing the scatterplot layout and coloring, and selection of
individual data points within the scatterplot updates individual and detail views to the right.
The individual NERO plot domain is the group orbit, and the interface permits moving within the
orbit to look at a particular transform of a single sample, with real-time updates of the model
output.
In the MNIST interface, for example, clicking and dragging within the polar plot selects and
changes the rotation angle, and updates the resulting rotated digit image and the models' outputs
from it.
Our interface is implemented in PySide (Python bindings for QT) as a desktop application,
running on the same machine as the model.

\subsection{Consensus}\label{subsec:consensus}
Although existing scalar metrics (accuracy, confidence) serve well as NERO metrics to incorporate
easier adaptions for practitioners, NERO evaluation should ideally also work on unlabeled data
lacking ground truth.
Because, as shown in Fig.~\ref{fig:general_illustration}, equivariance is revealed through the gap
between $h(x')$ and $y'$, which technically should not depend on the existence of ground truth.
However, given that existing metrics all require ground truth, an additional modest contribution
of this work is \textit{consensus}, which serves as a proxy for ground truth in the metric
evaluation, when making NERO evaluations for models with desired equivariance or covariance (as
opposed to invariance).
The consensus for input $x$ is roughly the average of the un-transformed model outputs on all
transformed inputs within the orbit.
Relative to Fig.~\ref{fig:general_illustration}, we have
\begin{equation} \label{eqn:consensus}
  \mathrm{consensus}(x) = \left\langle \tilde{\phi}(g^{-1},h(\phi(g,x))) \right\rangle_{g \in G}
\end{equation}
%
% The details of computing in output space $Y$ the average $\langle\cdot\rangle_G$ depend on the
% structure of $Y$, while $G$ depends on the equivariance of interest.
The average $\langle\cdot\rangle_G$ depends on the structure of output space $Y$, while $G$
depends on the equivariance of interest.
For object detections, $Y$ is the set of bounding boxes defined by corners
$(\mathrm{x}_\mathrm{min}, \mathrm{y}_\mathrm{min})$ and $(\mathrm{x}_\mathrm{max},
\mathrm{y}_\mathrm{max})$, and an element $(t_\mathrm{x}, t_\mathrm{y})$ of translation group $G$
acts on the bounding box by component-wise addition.
In this case, (\ref{eqn:consensus}) can be computed by simple arithmetic mean of the
translated bounding box corners.
%

%-------------------------------------------------------------------------
\section{Experiments - Applying NERO to More ML Cases}\label{sec:experiments}
We illustrated in \S\ref{sec:method} the designs and components of NERO evaluation through a 2D
digit classification task with MNIST~\cite{lecunGradientbasedLearningApplied1998}.
As mentioned before, NERO evaluation is model- and task-agnostic.
In this section, we demonstrate how NERO evaluation could be applied to different ML models in
three different research areas: object detection (classification and localization in 2D
photographic images), particle image velocimetry (velocity measurements in fluid dynamics),
and point clouds recognition (classification in 3D computer vision)
in \S\ref{subsec:object_detection}, \S\ref{subsec:piv} and
\S\ref{subsec:point_cloud_classification}, respectively.
In all subsections, we end with qualitative feedback from PhD students from our institution who
are knowledgeable in each area, but not involved in the developmet of NERO evaluation.

\subsection{Object Detection}\label{subsec:object_detection}
Object detection is a staple of computer vision research, witnessing dramatic advances from deep
learning~\cite{zhao_object_2019, deng_review_2020}.
Despite the great successes, recent research discovers that object detectors can be very vunerable
to small translations~\cite{10.1007/978-3-030-65414-6_4}.
And ongoing efforts have been dedicated on developing shift-equivariant neural
networks~\cite{chaman2021truly, 10.1007/978-3-030-65414-6_4}.
Unfortunately, evaluations for shift equivariance advancements are still based on the average of
scalar metrics such as precision, recall, intersection over union (IOU), or mean average precision
(mAP), over the dataset.
As noted in \S\ref{sec:introduction}, however, such summary evaluations give no direct insight
about equivariance, which is a natural concern for applications like autonomous driving requiring
high equivariance not just by average, but also in corner cases.
A person should be recognized as such regardless of their position with the image,
which implies an equivariance evaluation over the group of image translations.

For this section, we use the popular Faster R-CNN~\cite{ren_faster_rcnn} framework,
with the MSCOCO~\cite{linMicrosoftCOCOCommon2015} dataset, though the creation
and display of NERO plots for this task is independent of model or dataset.

\begin{figure}
    \centering
    \includegraphics[width=.9\linewidth]{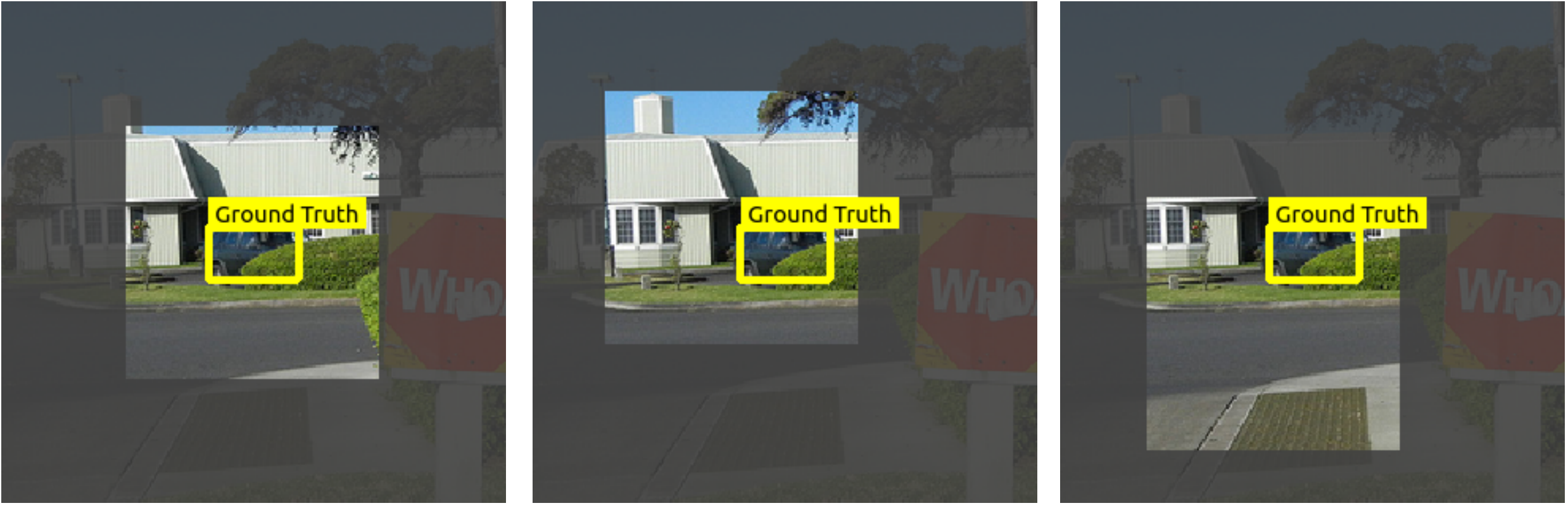}
    \caption{Key objects are shifted by cropping the original MSCOCO image to shifted bounds (the non-masked square).}
    \label{fig:coco_shift_examples}
    \vspace{-0.1in}
\end{figure}

\noindent\textbf{Data Preparation.}
The architecture of Faster R-CNN does not guarantee translational equivariance, so models with
different equivariance properties can be obtained by training with datasets with different
augmentations, as we show here.
We selected $5$ out of the $80$ MSCOCO classes for demonstration:
\textit{car}, \textit{bottle}, \textit{cup}, \textit{chair} and \textit{book}.
We selected objects that belong to these $5$ classes as key objects and cropped the original images
to a $128\times 128$ window around these objects.
As shown in Fig.~\ref{fig:coco_shift_examples}, translational shifts (by between $-64$ and $64$
pixels in both directions) are achieved by cropping with shifted bounds, so that the key object
positions change within the field of view.

To ensure interesting cropped images, the MSCOCO images are filtered with following criteria:
(1) include a key object whose ground truth class label is in the $5$
selected classes; (2) ensure that for all shifts the cropped fields of view does not extend
past the original image edges; and (3) ensure that the
key object's ground truth bounding box is not less than $1\%$ or more than $50\%$ of
the cropped $128\times 128$ region.

\noindent\textbf{Model Preparation.}
We predict that different levels of model equivariance can be created by different levels of
random shifts, or \textit{jittering}, in the training dataset of cropped images.
At $0\%$-jittering, key objects are never shifted and stay at the center of the cropped images for
training, while at $100\%$-jittering key objects are shifted randomly (uniformly) within the
$[-64, 64]$ range during training.
Jitterings are performed like other data augmentations, i.e., cropping happens at real-time in
data loaders.
A model trained with $0\%$ jittering is expected to only do well on unshifted images, while a
model from $100\%$ jittering is expected to be more equivariant (perform well regardless of shift).
\begin{figure*}
    \centering
    \includegraphics[width=.9\textwidth]{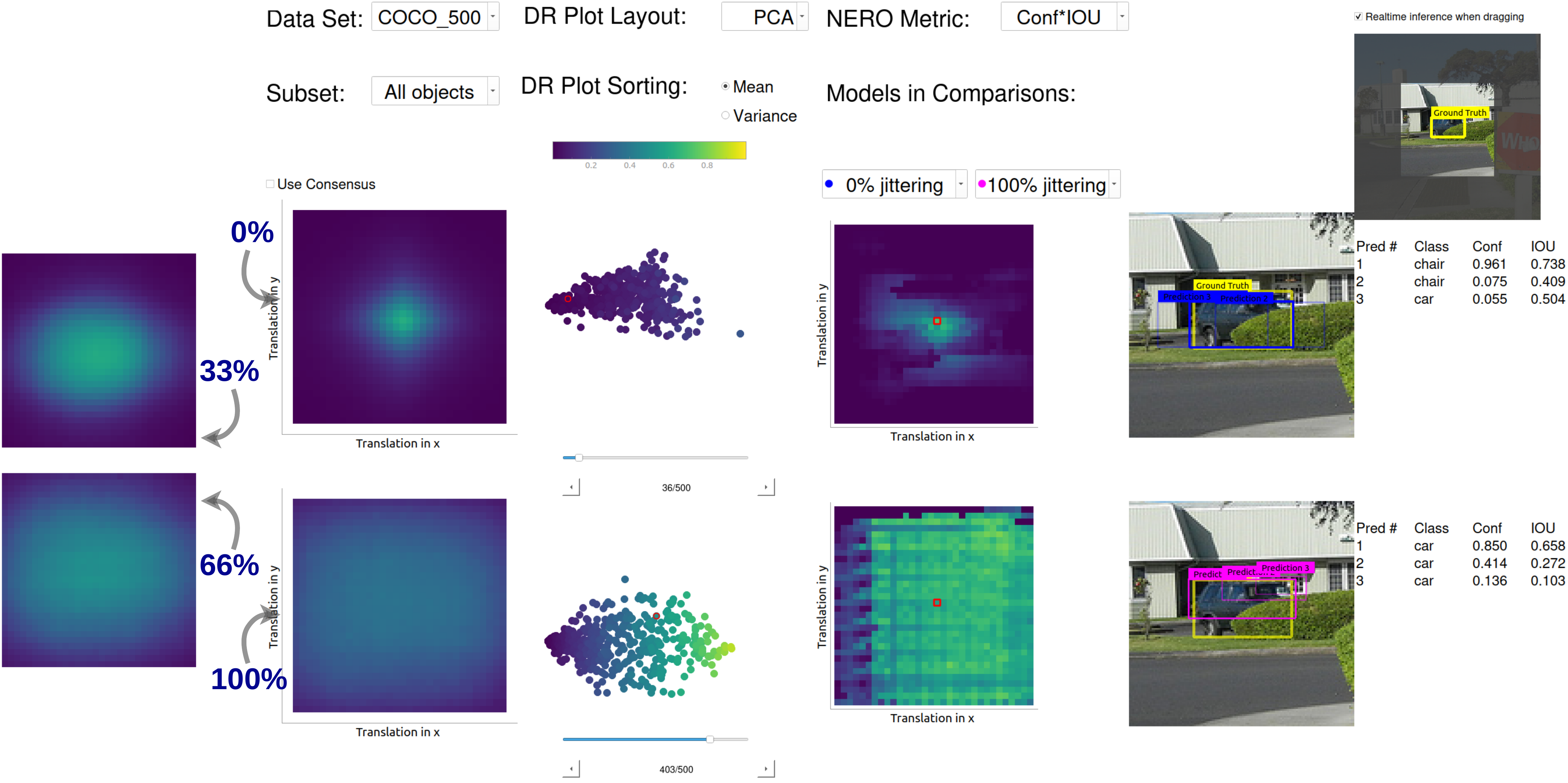}
    \caption{NERO interface for object detection, for models trained with
    $0\%$ (upper row) and $100\%$ (lower row) jittering. Sections for aggregate, dimension
    reduction, individual, and detail plots are organized as in the MNIST interface (Fig.~\ref{fig:mnist_interface}).
    Two aggregate NERO plots on left edge show intermediate jittering levels for comparison.}
    \label{fig:coco_interface}
    \vspace{-0.1in}
\end{figure*}

\noindent\textbf{Results.}
Fig.~\ref{fig:coco_interface} shows the full NERO interface for models with $0\%$ and $100\%$
jittering.
As in the MNIST example of Fig.~\ref{fig:mnist_interface}, equivariance of the two models is
evaluated and visualized with both aggregate and individual NERO plots, connected with dimension
reduction plots, with a different task-appropriate detail display on the right.
The left edge of Fig.~\ref{fig:coco_interface} also shows aggregate NERO plots from two other
intermediate jittering levels.
Matching our expectations, the amount of jittering is visually reflected in the width of the
NERO plot peak, with high equivariance in $100\%$ jittering (lower row) evident in the wide
uniform plateau of high values in that heatmap, versus the small bright spot at $0\%$ jittering
(upper row) indicating non-equivariance.

\begin{figure}
    \centering
    \includegraphics[width=\linewidth]{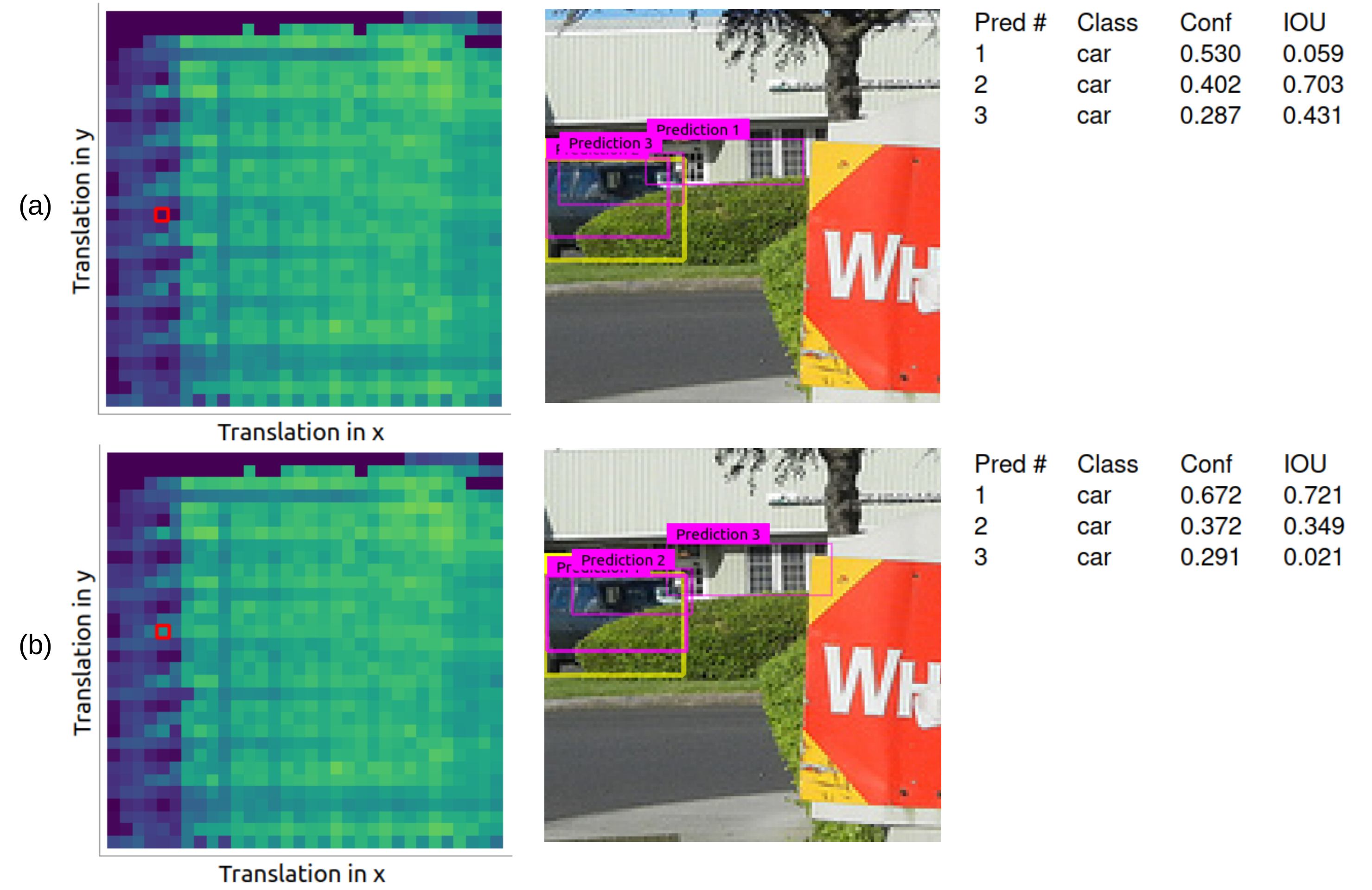}
    \caption{Individual NERO and detail plots further investigating model performance. (a):
    investigating $100\%$ jittering model's dark spot on its individual NERO plot. (b):
    investigating a nearby spot (input image similarly shifted) that has much better results of
    from the same $100\%$ jittering model.}
    \label{fig:coco_example}
    \vspace{-0.1in}
\end{figure}

Aggregate NERO plots give a quick overview of model equivariance, but individual NERO plots enable
detailed investigation.
For example, in Fig.~\ref{fig:coco_interface}, the individual NERO for the $100\%$ jittering model
has dark regions on the left edge, indicating worse performance at certain shifts.
A curious practitioner can simply click on those spots to scrutinize model details, as shown in
Fig.~\ref{fig:coco_example}, which investigates a small change in shift between the top and bottom
row.
We learn that at both shifts, the model gives three bounding box predictions, one with a high
IOU of about $0.7$, but the confidence ranking of the three boxes is different in the two
locations.
The individual NERO plot shows the IOU only for the most-confident prediction, creating the dark
regions.
In this way, NERO plots allow practitioner to explore and understand model edge cases.

\begin{figure}
    \centering
    \includegraphics[width=0.8\linewidth]{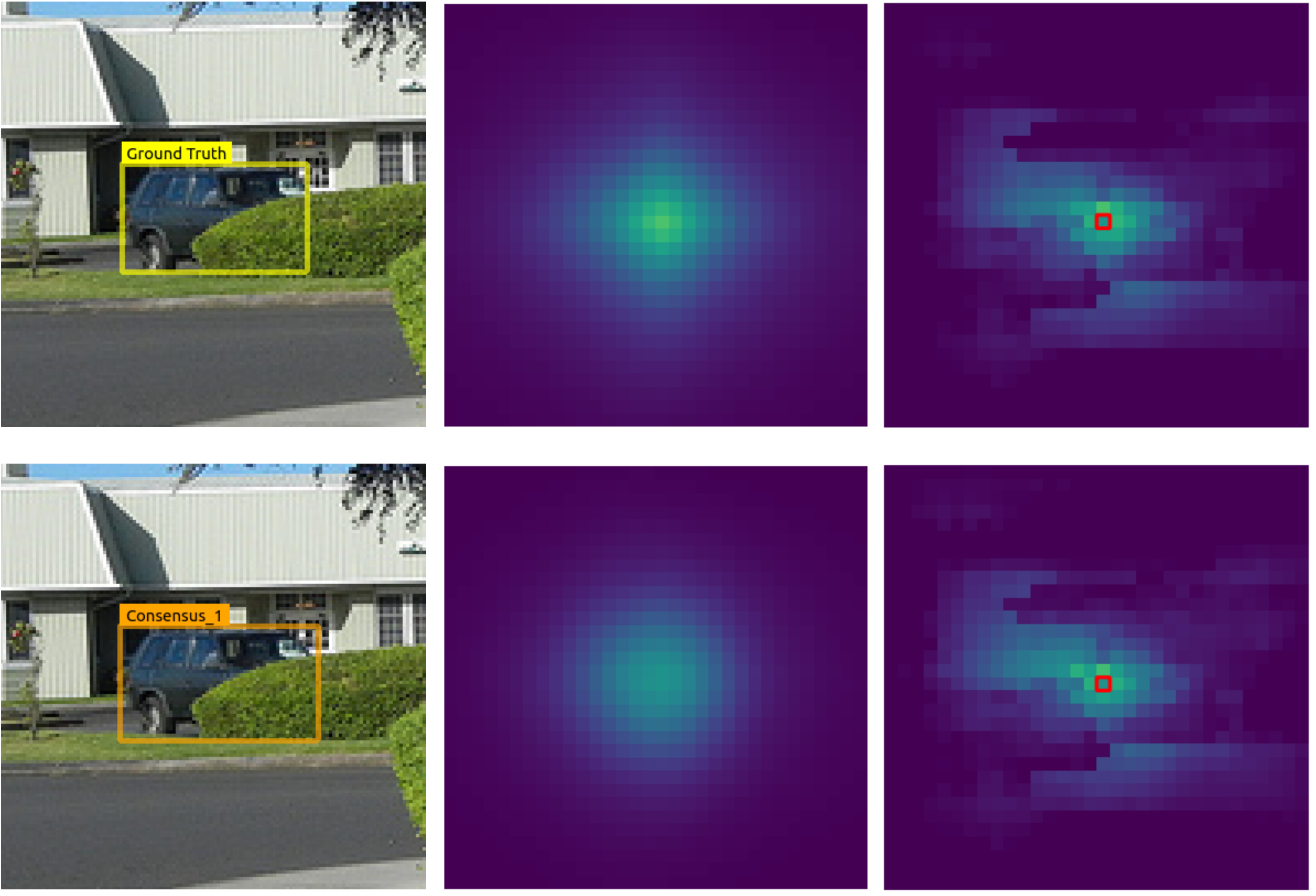}
    \caption{Consensus boxes computed from model outputs (left column), individual NERO plots of
    each model computed from ground truth (middle column) and from consensus (right
    column). Subtle differences are highlighted in yellow circles.}
    \label{fig:coco_consensus}
    \vspace{-0.1in}
\end{figure}

\noindent\textbf{Consensus.}
Consensus (\S\ref{subsec:consensus}) in this case is the average of unshifted bounding box
predictions from shifted input images.
Fig.~\ref{fig:coco_consensus} shows the individual NERO plots computed from ground truth and
consensus.
The strong similarity of the two plots suggest that, at least for this image, the amount and
structure of equivariance shown by NERO plots is nearly the same with or without ground truth,
which increases the applicability of NERO plots for unlabeled data.

\noindent\textbf{Expert Evaluation.}
A researcher in causality for ML, with basic knowledge of equivariance, tried our NERO interface
for object detection.
The evaluation was semi-guided, meaning that the expert was free to explore himself after we walked
him through examples similar to those earlier in this section.
The ensuing discussion focused on the NERO plot idea itself and its value; quotes below from the
expert are in italics.

\textit{It makes sense to compute equivariance like this, it is neat to model it with simple group
theories} -- the expert understood how we transform samples along group orbits, and measure
results on transformed samples.
\textit{Aggregate NERO plots are quick to look at when comparing two models, it's pretty much as
fast as looking at two numbers.} -- the expert felt that NERO plots do not create excessive visual
complexity for users.
\textit{I am surprised by how different the models can fail equivariance, and the fact that you
can just click on these dots to further explore is amazing ...} -- the expert said about the
DR plots -- \textit{... but now I see what are the differences} -- the expert looking at the
corresponding individual and detail plots.
Finally, after using the interface for about $10$ minutes, the expert concluded: \textit{Using
equivariance as an evaluation strategy is interesting and new, I've never thought about it, and I
am curious to see it with more examples. I think it is quite surprising to see these results, I
knew there is more going on underneath the average errors we see everyday, but being able to
actually see them and look further inside is amazing. I think it would benefit anyone who cares
about model equivariance or develops better ENN's}.

\subsection{Particle Image Velocimetry (PIV)}\label{subsec:piv}
Particle Image Velocimetry (PIV) is an important tool for physicists studying experimentally
constructed (as opposed to simulated) fluid dynamics.
PIV estimates velocity flow fields from frames of video of illuminated particles moving through a
flow domain.
Traditional PIV algorithms~\cite{westerweel_fundamentals_1997, heitz_variational_2010}
work for simple flows, but researchers are interested in the promise of ML-based methods for
faster computation and complex
flows~\cite{rabault_performing_2017, lee_piv_dcnn_2017, cai_dense_2019, cai_particle_2020}.

Empirical scientists expect measurement tools to respect physical symmetries.
For example, the world-space flow pattern recovered in PIV should be the same regardless of the
coordinate frame of measurement.
This corresponds to various rotations and reflections of the image data and covariant changes in
the recovered vector fields.
Continuous rotations are applicable here (as demonstrated on MNIST \S\ref{sec:method}),
but to demonstrate a different kind of NERO plot we worked with the dihedral group $D_4$ of
symmetries of the square (with reflections and $90^\circ$ rotations)~\cite{rotman2012introduction},
times the binary group $\mathbb{Z}_2$ for flow reversal, giving a discrete group of $16$ elements.
We investigate NERO plots for a PIV ML model to illustrate how evaluating model equivariance may
increase trust in novel scientific ML applications, by giving detailed information not captured in
existing summary ML evaluations (e.g. RMSE).
As PIV is closely related to the computer vision problem of recovering optical flow from video,
this example also suggests how NERO plots may work for optical flow models~\cite{hur_optical_2020}.

We compare a traditional (non-ML) Gunnar-Farneback method~\cite{gunnar_farneback}, implemented in
OpenCV~\cite{opencv_library}, with a recent deep learning method,
PIV-LiteFlowNet-en~\cite{cai_particle_2020}.
Training and test images are obtained from the Johns Hopkins Turbulence
Database~\cite{JHTDB_1, JHTDB_2}.

\noindent\textbf{Data Preparation.}
In total, $8,794$ pairs of images covering $6$ different types of flows, namely
\textit{Uniform}, \textit{Backstep}, \textit{Cylinder}, \textit{SQG}, \textit{DNS},
and \textit{Isotropic}, are used during training.
$120$ image pairs are used in testing when generating the NERO plots.
\begin{figure*}
    \centering
    \includegraphics[width=0.8\textwidth]{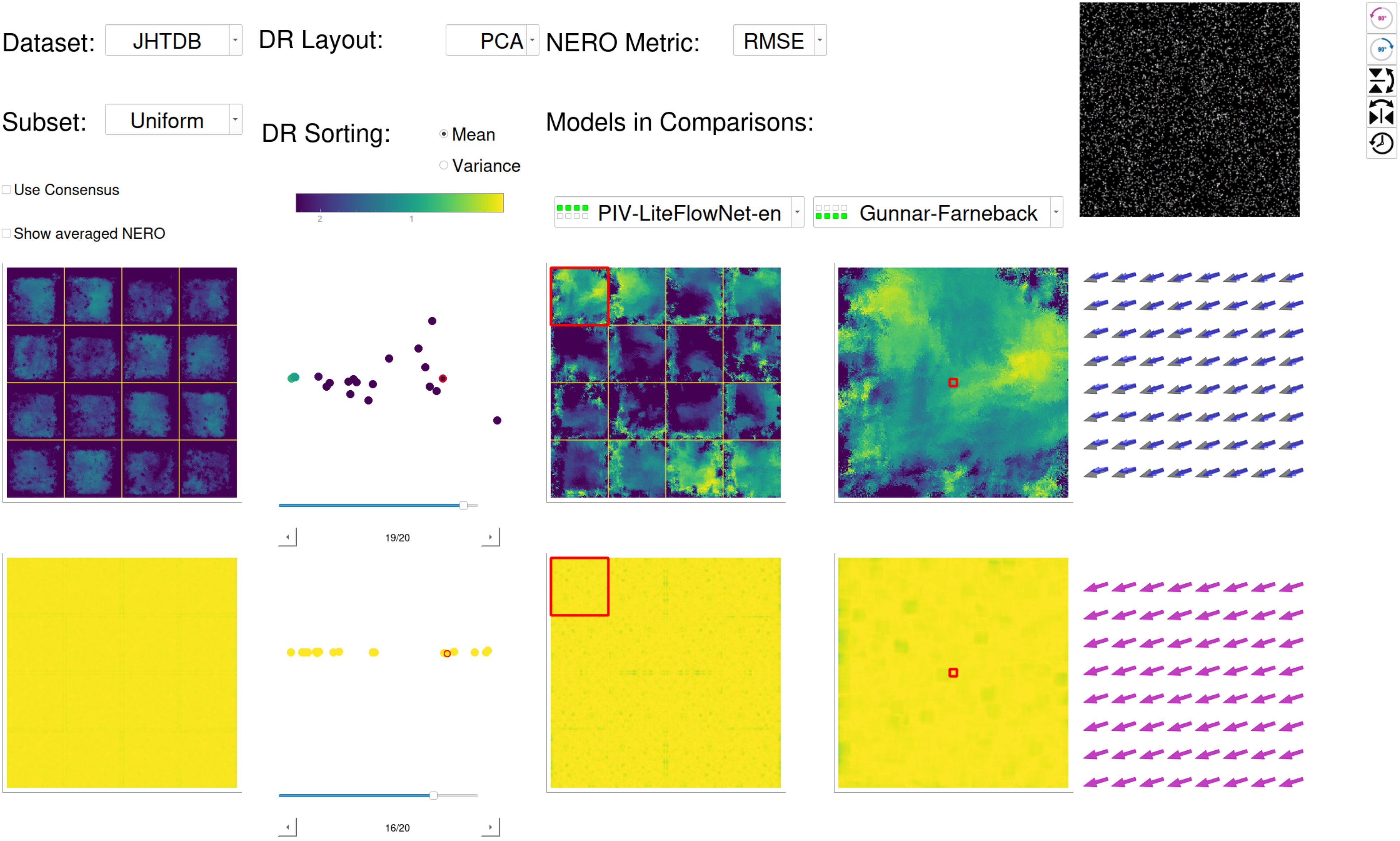}
    \caption{The NERO interface for PIV comparing an ML method (top row) with a non-ML method (bottom row).
    This has the same sections as in previous interface examples, but
    with a small-multiples display of the vector field domain for each element of the discrete orbit.}
    \label{fig:piv_interface}
    \vspace{-0.1in}
\end{figure*}

\noindent\textbf{Model Preparation.}
PIV-LiteFlowNet-en~\cite{cai_particle_2020} is trained with $8,794$ pairs of particle images
as explained above; Gunnar-Farneback does not require training.
Both are tested with the same test dataset consisting of $120$ image pairs.
Apart from performance as measured by RMSE, we expect Gunnar-Farneback to be naturally equivariant,
without bias towards any flow direction.
On the other hand, we expect less equivariance from PIV-LiteFlowNet-en, even though the training
and testing flow types are the same.

\begin{figure}
    \centering
    \begin{subfigure}[b]{0.325\linewidth}
        \centering
        \includegraphics[width=\textwidth]{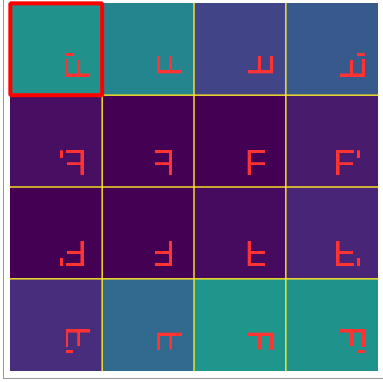}
        \caption{}
        \label{fig:piv_average}
    \end{subfigure}
    \begin{subfigure}[b]{0.325\linewidth}
        \includegraphics[width=\textwidth]{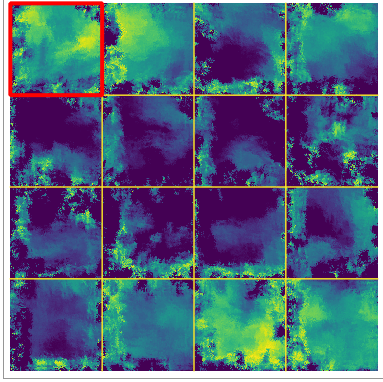}
        \caption{}
        \label{fig:piv_non_average}
    \end{subfigure}
    \caption{NERO plots for PIV with (a) spatially averaged error and (b) detailed display
    of the per-location error.}
    \label{fig:piv_average_example}
    \vspace{-0.1in}
\end{figure}

\noindent\textbf{Results.}
Fig.~\ref{fig:piv_interface} shows our NERO interface for comparing PIV-LiteFlowNet-en (top row)
and Gunnar-Farneback (bottom row).
The top right corner shows a controllable animation of the particle image sequence that PIV
analyzes.
As before, higher equivariance is shown with brighter and more uniform NERO heatmaps; dark spots
indicate non-equivariance.
Despite the similar use of a heatmap, NERO plots for PIV are richer than those used for object
detection (Figs.~\ref{fig:coco_interface}, \ref{fig:coco_example}).
In object detection NERO plots, each pixel represents one point in the orbit, i.e., a
specific shift.
Carrying the same idea to PIV would create NERO plots like Fig.~\ref{fig:piv_average}, with $16$
squares for each element of the (discrete) group orbit, as indicated with symbols in each bottom
right corner (\texttt{F} is original, \texttt{F'} is time-reversed, at all possible orientations),
with the heatmap showing RMSE over the whole flow domain.
Drilling down further, Figs.~\ref{fig:piv_non_average} and \ref{fig:piv_interface} use a
small-multiple display to show $16$ copies of the flow domain, to reveal the spatial locations of
flow for which the model was least equivariant.
The \texttt{Show averaged NERO} checkbox in the interface toggles between the two.

As expected, Fig.~\ref{fig:piv_interface} shows that Gunnar-Farneback performs consistently
better, with almost perfect equivariance.
To investigate further into model outputs, the individual detail plots include an enlarged view
of the non-averaged "pixel" in the individual NERO plot, and a vector glyph visualization overlays
the predicted field on the ground truth.

\noindent\textbf{Expert Evaluations.}
A physics researcher with expertise in PIV tried our NERO PIV interface and gave qualitative
feedback, in the same format as the previous object detection evaluation.
We walked our expert through the basic digit recognition task before letting him use the PIV
interface himself.
Quotes below from the expert are in italics.

\textit{It is very good to be able to see so much more information than an average value, which
tells too little of the story. You know, for a turbulence flow the interesting and hard part is not
everywhere, often much less than the boring part, so the average error really does not help much.}
-- the expert likes instantly that NERO plots show much richer information than conventional
scalar metrics.
\textit{I like that I am able to locate high-variance (less-equivariant) samples from the
dimension reduction plot, you see, this cluster has higher variance, and contains the interesting
samples} -- the expert said when looking at the DR plots -- \textit{it is really convenient to
select individual samples from the DR plots, it really brings out the interesting samples to
investigate} -- the expert thinks the design is effective in helping user traverse through samples
and locate the interesting one quickly.
\textit{Yes, definitely, it [NERO] would save me so much time analyzing PIV model outputs. I've
been using Jupyter notebook to visualize some parts of the flow but this [NERO] is totally a game
changer} -- the expert said when asked about if he would personally use the interface in his research.

\subsection{3D Point Cloud Classification}\label{subsec:point_cloud_classification}
Point cloud classification is a fundamental task in 3D computer vision that involves assigning
semantic labels to 3D point clouds~\cite{grilli2017review}.
It has many important applications, including object
recognitions~\cite{che2019object}, scene understanding~\cite{chen2019deep},
and robotics~\cite{duan2021robotics, wang2020grasping}.
Among many research areas in point cloud classification and analysis, equivariant point cloud
classification is a recent development that aims to battle the significant performance downgrade
caused by rotations by taking advantage of symmetry and invariance properties of
3D objects~\cite{chen2021equivariant, luo2022equivariant, finkelshtein2022simple}.
Current procedure on evaluating equivariance of these models is by augmenting random rotations to
test datasets and comparing average accuracies, as used in~\cite{chen2021equivariant}
and~\cite{luo2022equivariant}.
As discussed in \S\ref{subsec:enn}, such evaluation process has its drawback that models excelling
in the majority of transforms (rotations) will be good enough to excel in average metrics, without
real weakspots getting revealed, which hinders interpretability, trust and future developments.
In this section, we demonstrate how NERO evaluation mitigates this issue to support this
rising research area.

\begin{figure*}
    \centering
    \includegraphics[width=.9\textwidth]{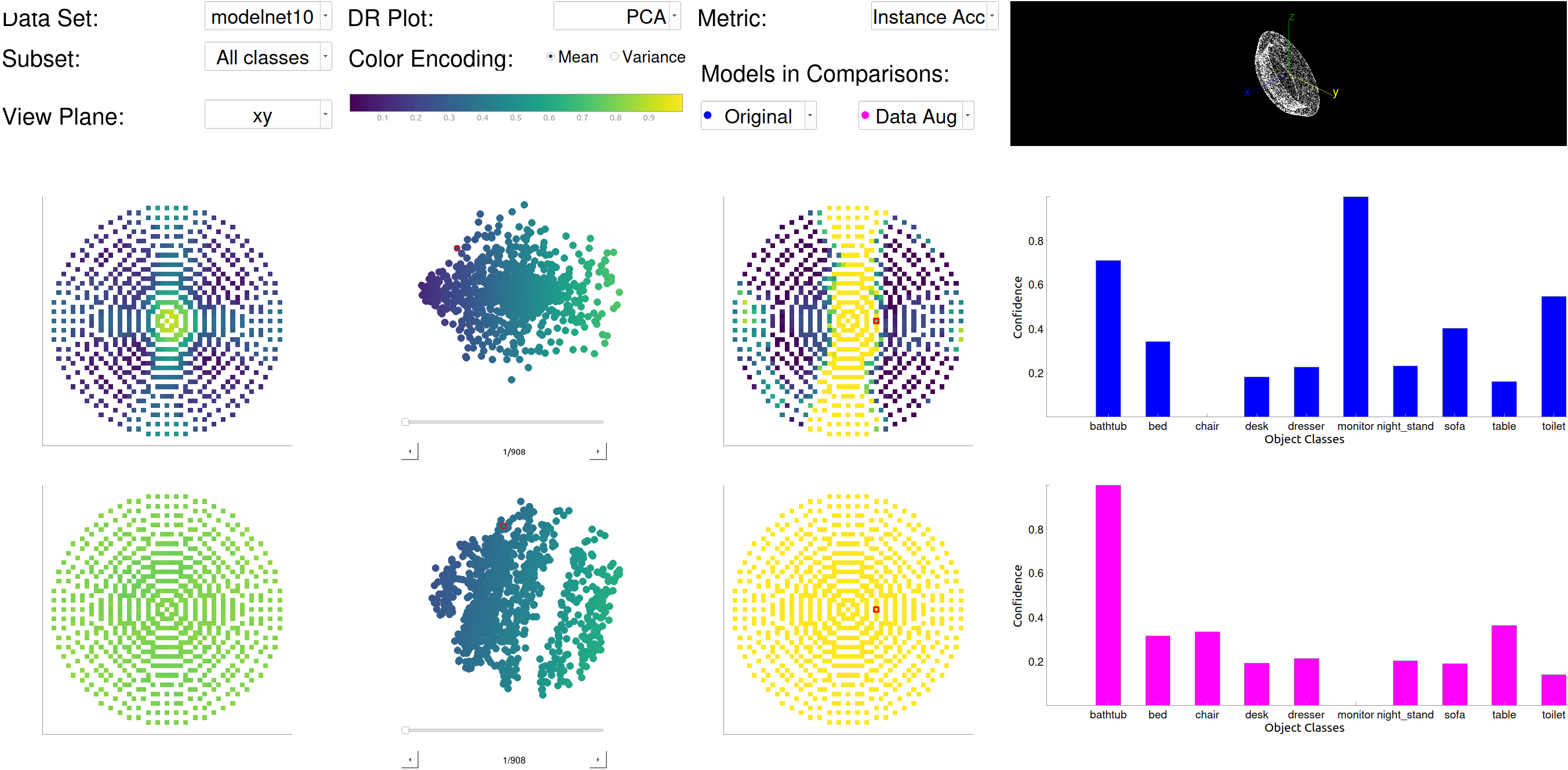}
    \caption{The NERO interface for 3D point cloud classification comparing Point Transformer model
    trained without (top row) and with (bottom row) rotation augmentations.
    This has the same sections as in previous interface examples.}
    \label{fig:point_cloud_interface}
\end{figure*}

To visualize results from 3D rotations in 2D NERO plots, we conduct our NERO evaluation based on a
subset of rotations.
More specifically, suppose each rotation is represented via an axis-angle representation, we
define each rotation axis to be a 3D vector sitting within one of the three 2D slicing planes,
namely \textit{x-y}, \textit{x-z}, and \textit{y-z} plane, with the vector's one end at the origin.
The angle in the axis-angle representation is simply a rotation angle between $0$ and $180$
degrees.
The angles between the rotation axis and its horizontal axis in the plane, alone with
the value of rotation angle, are visualized intuitively in a polar-coordinate plot, as shown in
both aggregate and individual NERO plots in Fig.~\ref{fig:point_cloud_interface}.

For this section, we demonstrate NERO evaluations on the Point Transformer~\cite{zhao2021point}
models with ModelNet40~\cite{7298801} dataset.

\noindent\textbf{Data Preparation.}
We use the widely adopted ModelNet40 and its subset ModelNet10 dataset~\cite{7298801}.
The ModelNet40 dataset contains $12,311$ CAD models with $40$ categories.
The dataset is split into $9,843$ training and $2,468$ testing samples.
ModelNet10 is a subset of ModelNet40 and only contains $10$ categories, with a total number of
$3,991$ training and $908$ testing samples.
As a common practice in point cloud classification, we follow the data preparation procedure from
Qi et. al.~\cite{10.5555/3295222.3295263} to uniformly sample point clouds from the CAD models.

\noindent\textbf{Model Preparation.}
When applying deep learning models on point cloud classifications, permutations of the point
clouds orderings is another common source of invariance besides rotations.
To make this demonstration more predictable, we exclude the effect from permutations
by choosing the Point Transformer~\cite{zhao2021point} model, which is by design invariant to
permutations thanks to its self-attention operator.
To show how NERO evaluations distinguish between a non- and equivaraint model, similar
to \S\ref{sec:method}, the Point Transformer model was trained twice, first without and then with
rotation augmentation, to create two models that differ predictably.
The augmentation model should have better invariance, even though the total amount of training
being the same.

\noindent\textbf{Results.}
Fig.~\ref{fig:point_cloud_interface} shows the NERO interface for the two models discussed above.
As in the MNIST example of Fig.~\ref{fig:mnist_interface}, invariances of the two models are
evaluated and visualized via aggregate and individual NERO plots, connected with dimension
reduction plots, with a different task-appropriate detail display on the right.
Looking at the aggregate plots on the left, we can observe that, matching our expectations, the
original model (top) has a bright spot at the center, indicating that it only performs well up
to small rotations (both axis and rotation angles), whilst the data-augmented model
(bottom) has a more uniform display across the plot, indicating that it is much more invariant
to rotations.

Individual NERO plots enable detailed investigations.
The specific sample shown in Fig.~\ref{fig:point_cloud_interface} is a bathtub.
And its default orientation in the dataset is vertical in terms of the $x$-$y$ plane.
From the bright yellow stripe in the top individual plot, we can observe that the original model is
only able to recover after rotations along vertical ($z$) axis, which are much easier,
while the augmented model (bottom) recognizes the bathtub across all axis-angle represented
rotations very well.

\noindent\textbf{Expert Evaluations.}
We invited the same expert from \S\ref{subsec:object_detection} to give us evaluations again.
This time, we focused more on collecting expert's feedbacks about how it feels going from one
interface to another.

\textit{It feels very similar from the object dection interface to this, the layouts are the same,
which is good, I am still able to quickly navigate myself to the places I am interested in}
-- the expert agrees that the identical high-level interface design helps researchers quickly
adapt from one application to another --
\textit{I probably have convered all my thoughts about using this interface in my previous
assessment, here is mostly the same, it is showing evaluation results way beyond scalar metrics,
and could be very useful when understanding and debugging model behaviors}
-- the expert agrees again that NERO evaluation provides more thorough and informative results
than standard scalar metrics.

As we have demonstrated how NERO evaluation can help better evaluate ML models as an integrated
workflow, we would like to point out that the reason of using data augmentations to generate ML
models is not to prove the effectiveness of data augmentation, but to generate models with
controllable behavior to prove and show the correctness as well as the benefits of the proposed
evaluation procedure.
%

%-------------------------------------------------------------------------
\section{Discussion, Conclusions, Future Work}\label{sec:discussion}
NERO evaluation is a new method of equivariance evaluation for ML models, built on basic group
theory, with an interactive user interface that we have demonstrated and evaluated in different
application settings.
The examples we have showed in Section~\ref{sec:method},~\ref{subsec:object_detection}
,~\ref{subsec:piv}, and~\ref{subsec:point_cloud_classification} demonstrate four settings where
NERO evaluations reveal model equivariance, better assess model performance, and make black-box
models more interpretable.
The idea of aggregate, dimension reduction, and individual NERO plots, linked in an interactive
interface, should in principle work in many other areas of applied ML research, to facilitate
finding and exploring model weak spots.

However, creating the NERO interface for a new area involves identifying a relevant and meaningful
transform group, designing a spatial layout of the group (the NERO plot domain), and creating
informative displays of the individual and detail views.
This has been largely straight-forward for our work to date, because the groups we have considered
have natural 2D layouts, being pure rotations
(\S\ref{sec:method},~\S\ref{subsec:point_cloud_classification}),
shifts (\S\ref{subsec:object_detection}),
or a combination of discrete rotations and flips (\S\ref{subsec:piv}).
For more abstract groups, like the symmetry group of permutations of $N$ nodes in a graph,
more careful and creative design is required.
Other spatial transform groups may be easier to address, such as \textit{scaling} image data.
We have not yet studied integrating this 1D transform group into our designs,
but are motivated to do so given its role in other equivariant neural network (ENN)
research~\cite{xuScaleInvariantConvolutionalNeural2014, kanazawaLocallyScaleInvariantConvolutional2014, worrallDeepScalespacesEquivariance2019, ghoshScaleSteerableFilters2019, sosnovikScaleEquivariantSteerableNetworks2020}.

We plan to  further studying the
idea of \textit{consensus} (\S\ref{subsec:consensus}, Fig.~\ref{fig:coco_consensus}), as it
potentially frees ML evaluation from needing ground truth.
We have not however quantitatively studied its reliability, nor characterized how much it assumes
the model being evaluated is already basically working (not generating noise).
Defining how to compute consensus in other more complicated domains is future work.

Adversarial attacks have been empirically shown to be devastating to neural networks performance
with pixel-wise~\cite{pixelwise_attack} perturbations and geometric
transformations~\cite{geo_attack_1,geo_attack_2} applied to the input data.
The vulnerability of neural networks could be further studied and analyzed with NERO evaluations.
More specifically, the relationship between equivariance and robustness has not been
scrutinized with an evaluation and visualization tool such as the one we proposed, which hinders
researching adversarial attacks through an equivariance lens.

As we briefly mentioned in \S\ref{subsec:iml}, NERO plots can be a drop-in replacement for the
conventional scalar-based evaluations that are widely employed in current ENN and surrogate models
studies.
In ENN, NERO plots provide a more thorough and direct comparisons between the existing models and
the new, more equivariant models.
And for surrogate models, since the explanation is based on the assumption that the surrogate
exhibits similar behavior as the original (to be explained) model, NERO evaluation may give a
stronger proof than existing summary scalar metrics.
Other future work includes conducting a more thorough user surveys, developing a web-based
interface for easier access, and creating a larger library of interface designs for
common ML applications.

\bibliographystyle{abbrv-doi-hyperref}
\newpage
\bibliography{reference}

%% ^^^^^   FOR IEEE VIS, EVERYTHING HERE MAY BE INCLUDED IN THE    ^^^^^ %%
%% 2-PAGE ALLOTMENT FOR REFERENCES, FIGURE CREDITS, AND ACKNOWLEDGEMENTS %%

% \appendix % You can use the `hideappendix` class option to skip everything after \appendix

% \section{About Appendices}
% Refer to \cref{sec:appendices_inst} for instructions regarding appendices.

% \section{Troubleshooting}
% \label{appendix:troubleshooting}

% \subsection{ifpdf error}

% If you receive compilation errors along the lines of \texttt{Package ifpdf Error: Name clash, \textbackslash ifpdf is already defined} then please add a new line \verb|\let\ifpdf\relax| right after the \verb|\documentclass[journal]{vgtc}| call.
% Note that your error is due to packages you use that define \verb|\ifpdf| which is obsolete (the result is that \verb|\ifpdf| is defined twice); these packages should be changed to use \verb|ifpdf| package instead.

% \subsection{\texttt{pdfendlink} error}

% Occasionally (for some \LaTeX\ distributions) this hyper-linked bib\TeX\ style may lead to \textbf{compilation errors} (\texttt{pdfendlink ended up in different nesting level ...}) if a reference entry is broken across two pages (due to a bug in \verb|hyperref|).
% In this case, make sure you have the latest version of the \verb|hyperref| package (i.e.\ update your \LaTeX\ installation/packages) or, alternatively, revert back to \verb|\bibliographystyle{abbrv-doi}| (at the expense of removing hyperlinks from the bibliography) and try \verb|\bibliographystyle{abbrv-doi-hyperref}| again after some more editing.

\end{document}